\documentclass[twoside]{article}

\usepackage{blkarray}
\usepackage{graphicx}
\usepackage{epsfig,times,subfigure,fullpage}
\usepackage{amssymb,amsopn,float,enumerate}
\usepackage{amssymb,algorithmicx,algpseudocode,algorithm,amsmath,relsize}
\usepackage{color}

\usepackage{amsopn,bm}
\usepackage{epstopdf}
\allowdisplaybreaks
\makeatletter
\g@addto@macro\normalsize{%
  \setlength\abovedisplayskip{2pt}
  \setlength\belowdisplayskip{2pt}
  \setlength\abovedisplayshortskip{2pt}
  \setlength\belowdisplayshortskip{2pt}
}
\newtheorem{theorem}{Theorem}
\newtheorem{Lemma}[theorem]{Lemma}
\newtheorem{proposition}[theorem]{Proposition}
\newtheorem{myDef}{Definition}[section]

\newcommand{\given}{\, | \,}
\newcommand{\cN}{{\cal N}}
\renewcommand{\u}{\mathbf{u}}

\newcommand\I{\mathbb{I}}
\newcommand{\ud}{\,\mathrm{d}}
\newcommand{\beq}{\begin{equation}}
\newcommand{\eeq}{\end{equation}}
\newcommand{\myref}[1]{(\ref{#1})}
\DeclareMathOperator{\Tr}{Tr}
\newcommand{\proof}{\noindent{\itshape Proof:}\hspace*{1em}}
\newcommand{\qed}{\nolinebreak[1]~~~\hspace*{\fill} \rule{5pt}{5pt}\vspace*{\parskip}\vspace*{1ex}}

\usepackage{times}

 \title{The Matrix Generalized Inverse Gaussian Distribution: \\ Properties and Applications}
\author{Farideh Fazayeli \qquad Arindam Banerjee  \vspace*{2mm}
\\
\{farideh,banerjee@cs.umn.edu\}\vspace*{2mm}\\
Department of Computer Science \& Engineering\\
University of Minnesota, Twin Cities}
\date{}
\begin{document}
\maketitle
\begin{abstract}
While the Matrix Generalized Inverse Gaussian ($\mathcal{MGIG}$) distribution arises naturally in some settings as a distribution over symmetric positive semi-definite matrices, certain key properties of the distribution and effective ways of sampling from the distribution have not been carefully studied.
In this paper, we show that the $\mathcal{MGIG}$ is unimodal, and the mode can be obtained by solving an Algebraic Riccati Equation (ARE) equation~\cite{boba91}. Based on the property, we propose an importance sampling method for the $\mathcal{MGIG}$ where the mode of the proposal distribution matches that of the target. The proposed sampling method is more efficient than existing approaches~\cite{yalz13,yoshii13}, which use proposal distributions that may have the mode far from the $\mathcal{MGIG}$'s mode.
%The current inference algorithm for mean of the $\mathcal{MGIG}$ ~\cite{yalz13,yoshii13} is not very effective since it use a proposal distribution that is very discrepant from the $\mathcal{MGIG}$.
Further, we illustrate that the the posterior distribution in latent factor models, such as probabilistic matrix factorization (PMF)~\cite{samn07}, when marginalized over one latent factor has the $\mathcal{MGIG}$ distribution. The characterization leads to a novel Collapsed Monte Carlo (CMC) inference algorithm for such latent factor models.
We illustrate that CMC has a lower log loss or perplexity than MCMC, and needs fewer samples.

%Probabilistic matrix factorization (PMF) and variants constitute arguably the most widely used approach to matrix completion. Existing inference algorithms for such models include point estimates by alternating minimization and Markov Chain Monte Carlo (MCMC) methods for Bayesian inference. The former are computationally efficient, but do not characterize uncertainties associated with the predictions and can have local minima issues. The latter have desirable asymptotic properties but can be slow in practice.
%In this paper, we show that one of the latent factors in PMF can be marginalized or collapsed analytically yielding a Matrix Generalized Inverse Gaussian ($\mathcal{MGIG}$) posterior distribution over the other latent factor. The characterization leads to a novel Collapsed Monte Carlo (CMC, not MCMC) inference algorithm for such models.
%We compare the proposed CMC with MCMC on Bayesian PMF using both synthetic and real datasets. We illustrate that CMC has a lower log loss or perplexity, need fewer samples, and can be substantially faster compared to MCMC.
\end{abstract}

\section{Introduction}
\label{sec:intro}
%importance of MGIG and why importance sampling
Matrix Generalized Inverse Gaussian ($\mathcal{MGIG}$) distributions \cite{babj82,butl98} are a flexible family of distributions over the space of symmetric positive definite matrices and has been recently applied as the prior for covariance matrix~\cite{li13,yalz13,yoshii13}. $\mathcal{MGIG}$ is a flexible prior since it contains Wishart, and Inverse Wishart distributions as special cases. We anticipate the usage of $\mathcal{MGIG}$ as prior for statistical machine learning models to grow with potential applications in Bayesian dimensionality reduction and Bayesian matrix completion. We illustrate some of these connections in Section \ref{sec:application}.

Some properties of the $\mathcal{MGIG}$ distribution and its connection with Wishart distribution has been studied in~\cite{butl98,seshadri03,seshadri08}.~However, to best of our knowledge, it is not yet known if the distribution is unimodal and, if it is unimodal, how to obtain the mode of $\mathcal{MGIG}$.~Besides, it is difficult to analytically calculate mean of the distribution and sample from the $\mathcal{MGIG}$ distribution.~Monte Carlo methods like the importance sampling can in principle be applied to infer the mean of $\mathcal{MGIG}$ but one needs to design a suitable proposal distribution~\cite{mackay03,owen13}.

\begin{figure}
\centering
\includegraphics[width=0.35\textwidth]{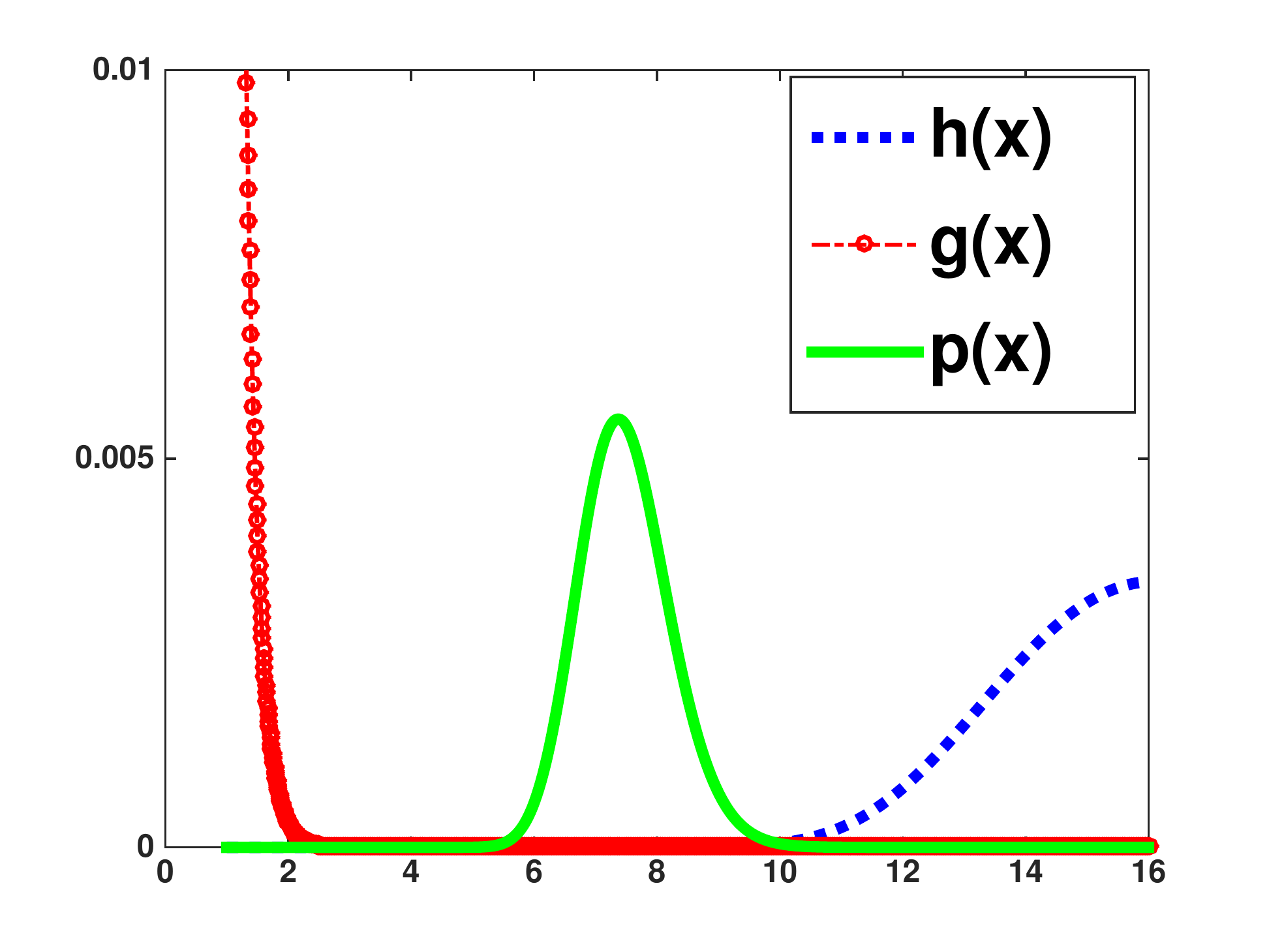}
\caption{An illustration of bad proposal distribution in importance sampling. Let $p(x) = h^*(x)g^*(x)/Z_p \propto h(x) g(x)$. Note that $h(x)=h^*(x)/Z_h$ nor $g(x)=g^*(x)/Z_g$ need not be a good candidate proposal distribution to approximate $p(x)$ since the mode of both $h(x)$ and $g(x)$ distribution is far away from $p(x)$.}
\label{fig:isschem}
\end{figure}

There is only one important sampling procedure for estimating the mean of $\mathcal{MGIG}$ \cite{yalz13,yoshii13}. In this approach, $\mathcal{MGIG}$ is viewed as a product of the Wishart and Inverse Wishart distributions and one of them is used as the proposal distribution.
However, we illustrate that the mode of the proposal distribution in \cite{yalz13,yoshii13} may be far away from the $\mathcal{MGIG}$'s mode. As a result, the proposal density is small in a region where the $\mathcal{MGIG}$ density is large yielding to an ineffective sampler and drastically wrong estimate of the mean (Figures \ref{fig:isschem} and \ref{fig:MGIGvsWIW}).

In this paper, %we provide a new important sampling technique for the $\mathcal{MGIG}$. 
we first illustrate that the $\mathcal{MGIG}$ distribution is unimodal where the  mode can be obtained by solving an {\em Algebraic Riccati Equation (ARE) \cite{boba91}}. This characterization leads to an effective importance sampler for the $\mathcal{MGIG}$ distribution. More specifically, for estimating the expectation $\mathbb{E}_{X\sim \mathcal{MGIG}}[g(X)]$, we select a proposal distribution over space of symmetric positive definite matrices like Wishart or Inverse Wishart distribution such that the mode of the proposal matches the mode of the $\mathcal{MGIG}$.~As a result, unlike the current sampler~\cite{yalz13,yoshii13}, by aligning the shape of the proposal and the $\mathcal{MGIG}$, the density of the proposal gets higher values in the high density regions of the target, yielding to a good approximation of $\mathbb{E}_{X\sim \mathcal{MGIG}}[g(X)]$.

Further, we discuss a new application of the $\mathcal{MGIG}$ distribution in latent factor models such as probabilistic matrix factorization (PMF)~\cite{samn07} or Bayesian PCA (BPCA)~\cite{bish99a} . In these settings, the given matrix $X \in \mathbb{R}^{N \times M}$ is approximated by a low-rank matrix $\hat{X} = U V^T$ where $U \in \mathbb{R}^{N \times D} $ and $V \in \mathbb{R}^{M \times D}$ with Gaussian priors over the latent matrices $U$ and $V$.
We show that after analytically marginalizing one of the latent matrices in PMF (or BPCA), the posterior over the other matrix has the $\mathcal{MGIG}$ distribution.  
This illustration yields to a novel Collapsed Monte Carlo (CMC) inference algorithm for PMF. In particular, we marginalize one of the latent matrices, say $V$, and propose a direct Monte Carlo sampling from the posterior of the other matrix, say $U$. %, by using the importance sampling procedure for the $\mathcal{MGIG}$.
Through extensive experimental analysis on synthetic, SNP, gene expression, and
MovieLens datasets, we show that CMC has lower log loss or perplexity with fewer samples than Markov Chain Monte Carlo (MCMC) inference approach for PMF~\cite{samn08b}. 

The rest of the paper is organized as follows. In Section \ref{sec:back}, we cover background materials. In Section \ref{sec:mgig}, we show that $\mathcal{MGIG}$ is unimodal and give a novel importance sampler for $\mathcal{MGIG}$.
 We provide the connection of $\mathcal{MGIG}$ with PMF in Section \ref{sec:application}, present the results in Section \ref{sec:res}, and conclude in Section \ref{sec:concl}.

\section{Background and Preliminary}
\label{sec:back}
%In this Section, we give a brief overview of $\mathcal{MGIG}$ distribution, Algebraic Riccati Equations (ARE), and PMF. and introduce some tools that will be used in our analysis.
In this section we provide some background on the relevant topics and tools that will be used in our analysis. We start by an introduction to importance sampling, $\mathcal{MGIG}$ distribution, a brief overview of the algebraic Riccati Equations (ARE), followed by describing the connection between Probabilistic Matrix Factorization and PCA.

{\bf{Notations:}} Let $\mathbb{S}^N_{++}$ and $\mathbb{S}^N_{+}$ denote the space of symmetric ($N \times N$) positive definite and positive semi-definite matrix, respectively. Denote $0_N$ as a matrix of all zeros $\{0\}^{N \times N}$. Let $|.|$ denote the determinant of matrix, $\Tr(.)$ be the matrix trace. A matrix random variable $\Lambda \in \mathbb{S}^N_{++}$ has a Wishart distribution and is denoted as $\mathcal{W}_N(\Lambda | \Phi, \tau)$ where $\tau > N-1$ and $\Phi \in \mathbb{S}^N_{++}$~\cite{wish28}. A matrix random variable $\Lambda \in \mathbb{S}^N_{++}$ has an Inverse Wishart distribution and is denoted as $\mathcal{IW}_N(\Lambda | \Psi, \alpha)$ where $\alpha > N-1$ and $\Psi \in \mathbb{S}^N_{++}$ is the scale matrix. Consider the matrix $X \in \mathbb{R}^{N \times M}$. We denote ${\bf{x}}_{:m}$ as the $m^{th}$ column of $X$ and ${\bf{x}}_n$ as the $n^{th}$ row of $X$.

\subsection{Importance Sampling}
Consider distribution $p(x) = \frac{1}{Z_p} p^*(x)$ where $Z_p$ is the partition function which plays the role of a normalizing constant.
Importance sampling is a general technique for estimating $\mathbb{E}_{x\sim p(x)}[f(x)]$ where sampling from $p(x)$ (the target distribution) is difficult but we can evaluate the value of $p^*(x)$ at any given $x$ \cite{mackay03}. The idea is to draw $S$ samples $\{x_i \}_{i=1}^S$ from a similar but easier distribution  denoted by proposal distribution $q(x) = \frac{1}{Z_q} q^*(x)$ and calculate the expected value as follows
\begin{eqnarray}
\mathbb{E}_{x\sim p}[f(x)] = \mathbb{E}_{x \sim q} \left[\frac{f(x) p(x)}{q(x)} \right]
\approx   \frac{\sum_{i=1}^S f(x_i) w(x_i)}{\sum_{i=1}^S w(x_i)}, \nonumber
\label{eq:importSam}
\end{eqnarray}
where $w(x_i) = \frac{p^*(x_i)}{q^*(x_i)}$ is the weight of each sample $i$, and ${Z_q}$ is the partition function. 

%The first step in the IS scheme is to define a proposal distribution that is similar to the target distribution and easier to sample from.
The efficiency of importance sampling depends on how closely the proposal approximates the target in the shape. One way for monitoring the efficiency of importance sampling is the effective sample size measured as
$ESS = \frac{(\sum_{i=1}^S w(x_i))^2}{\sum_{i=1}^S w^2(x_i)}$
\cite{kong94}.
%The $ESS$ takes values between 1 and $S$. 
If the proposal has the same shape as the target distribution, $ESS$ achieves the maximum value. The other extreme happens if all but one of the importance
weights are zero yielding to a minimum $ESS$ value of one.~Very small value of $ESS$ indicates a big discrepancy between the proposal and target (for example when the mode of the proposal distribution is far away from the target's mode) leading to a drastically wrong estimate of $\mathbb{E}_{x\sim p}[f(x)]$~\cite{mackay03}.

In particular, consider target distributions that can be decomposed as a product of two distributions, i.e., $p(x) \propto h(x) g(x)$. Naturally, one may choose one of the multiplicand of $p(x)$ ($h(x)$ or $g(x)$) as the proposal distribution since weight calculation becomes the evaluation at the other multiplicand. 
However, when mode of $h(x)$ or $g(x)$ is far away from the $p(x)$'s mode, neither of them are an appropriate candidate for the proposal distribution since the proposal density is small in a region where the target density is large (Figure \ref{fig:isschem}). Thus it is quite possible that the sampler has a very low effective sample size ($ESS$)~\cite{kong94}.

%For example, while the prior has been chosen as the proposal distribution for estimating mean of the posterior~\cite{fung89,shachter89}, it might yield to an inefficient sampler
%if the data are very informative and the posterior is very different from prior. To resolve this issue, proposal distribution based on mode matching is used to infer the mean of posterior distribution~\cite{owen13}.

\subsection{$\boldsymbol{\mathcal{MGIG}}$ Distribution}
$\mathcal{MGIG}$ distribution was first introduced in~\cite{babj82} as a distribution over the space of symmetric ($N \times N$) positive definite matrices defined as follows.
\begin{myDef}[$\boldsymbol{\mathcal{MGIG}}$ Distribution]
\label{def:MGIGdens}
Let $\Lambda$ be a symmetric ($N \times N$) positive definite matrix.
A matrix-variate random variable $\Lambda$ is $\mathcal{MGIG}$  distributed \cite{babj82,butl98} and is denoted as  $\Lambda \sim \mathcal{MGIG}_N(\Psi, \Phi, \nu)$ if the density of $\Lambda$ is
\begin{eqnarray}
f(\Lambda) = \frac{ \mid \Lambda \mid ^{ \nu - (N+1)/2} } { \mid \frac{\Psi}{2} \mid ^ {\nu} ~ B_\nu (\frac{\Phi}{2}\frac{\Psi}{2}) } \exp \{\Tr( -\frac{1}{2} \Psi \Lambda^{-1} -\frac{1}{2} \Phi \Lambda  ) \}, \nonumber
\label{eq:MGIGdens}
\end{eqnarray}
where $B_\nu(.)$ is the matrix Bessel function \cite{herz55} % as defined in Definition \ref{def:bessel}.
 defined as
\begin{align}
B_\nu(\frac{\Phi}{2}\frac{\Psi}{2}) = |\frac{\Phi}{2}|^{-\nu} \int_{\mathbb{S}^N_{++}} |S|^{-\nu - \frac{N+1}{2}} \exp \{\Tr( - \frac{1}{2} \Psi  S^{-1} -\frac{1}{2} \Phi S ) \} dS.
\label{eq:bessel}
\end{align}
% %and
% %Barndorff-Nielsen et al., 1982; Zhang et al., 2012) and is formally proposed by Butler (1998).
%The domain for parameters $\Phi$ and $\Psi$ for $N \geq 2$ is
% \begin{align}
% & \{ \Psi \in \mathbb{S}^N_{+} , \Phi  \in \mathbb{S}^N_{++}  \} & ~ \textrm{if} ~ &  \nu \geq \frac{1}{2}N, \notag \\
% & \{ \Psi \in \mathbb{S}^N_{++}, \Phi  \in \mathbb{S}^N_{++}  \} & ~ \textrm{if} ~  & -\frac{1}{2}(N-1) \leq \nu < \frac{1}{2}N, \notag \\
% & \{ \Psi\in \mathbb{S}^N_{++}, \Phi \in \mathbb{S}^N_{+}  \} & ~ \textrm{if} ~ & \nu < -\frac{1}{2}(N-1)~. \notag
% \end{align}
\end{myDef}

%Next, we discuss special cases of $\mathcal{MGIG}$ distribution.
When $N=1$, the $\mathcal{MGIG}$ is  the generalized inverse Gaussian distribution $\mathcal{GIG}$~\cite{jorgensen82} which is often used as the prior in several domains~\cite{blei10,eberlein95}. In the following, we show that if $\Psi=0$, the $\mathcal{MGIG}$ distribution reduces to the Wishart, and if $\Phi=0$, it becomes the Inverse Wishart distribution.

\begin{proposition}\cite[Proposition 2]{yalz13}
 If matrix $\Lambda \sim \mathcal{MGIG}_N(\Psi, \Phi, \nu)$, then $\Lambda^{-1} \sim \mathcal{MGIG}_N(\Phi, \Psi, -\nu)$.
 \label{prop:invMGIG}
\end{proposition}
\proof
The proof follows from the Bessel function property $B_\delta(WZ) = |WZ|^{-\delta} B_{-\delta}(ZW)$ \cite{yalz13}.
\qed

\begin{proposition}
If matrix $\Lambda \sim \mathcal{MGIG}_N(\Psi, 0_N, \nu)$, and $-\nu >  \frac{N-1}{2}$, then $\Lambda \sim IW_N(\Psi, -2\nu)$.
\label{prop:IWMGIG}
\end{proposition}
\proof
First note that If $-\nu > \frac{N-1}{2}$,  then we have $
B_{\nu}(0_N) = \Gamma_N(-\nu)$ \cite{buwo03}.
Then, the proof simply follows from Definition \ref{def:MGIGdens}.
%% and Definition \ref{def:invwishart}.
%From Definition \ref{def:MGIGdens}, the density of $\Lambda$ is
% \begin{align}
% f(\Lambda) &= \frac{ \mid \Lambda \mid ^{ \nu - (N+1)/2} } { \mid \frac{\Psi}{2} \mid ^ {\nu} ~ B_\nu (0_N) } \exp \{\Tr( -\frac{1}{2} \Psi \Lambda^{-1}) \} \\
% &=\frac{ \mid \Lambda \mid ^{ \nu - (N+1)/2} } { \mid \frac{\Psi}{2} \mid ^ {\nu} ~ \Gamma_N(-\nu) } \exp \{\Tr( -\frac{1}{2} \Psi \Lambda^{-1}) \},
% \end{align}
%which is the density function of $\Lambda \sim IW_N(\Psi, -2\nu)$.
%This completes the proof.
\qed

\begin{proposition}
If matrix $\Lambda \sim \mathcal{MGIG}_N(0_N, \Phi, \nu)$, and $\nu >  \frac{N-1}{2}$, then $\Lambda \sim W_N(\Phi^{-1}, 2\nu)$.
\end{proposition}
\proof
From Proposition \ref{prop:invMGIG}, we have $\Lambda^{-1} \sim \mathcal{MGIG}_N(\Phi, 0_N, -\nu)$.
Also, from Proposition \ref{prop:IWMGIG}, we have $\Lambda^{-1} \sim IW_N(\Phi, 2\nu)$.
If matrix $\Lambda^{-1} \sim IW_N(\Phi, 2\nu)$ then $\Lambda \sim W_N(\Phi^{-1}, 2\nu)$.
This completes the proof.
\qed
\vspace{0.2cm}
\noindent{\bf{Sampling Mean of $\boldsymbol{\mathcal{MGIG}}$:}} The sufficient statistics of $\mathcal{MGIG}$ are $\log |\Lambda|$, $\Lambda$, and $\Lambda^{-1}$. It is, however,
difficult to analytically calculate the expectations $\mathbb{E}_{\Lambda \sim\mathcal{MGIG}}[\Lambda]$ and $\mathbb{E}_{\Lambda \sim\mathcal{MGIG}}[\Lambda^{-1}]$. 
Importance sampling can be applied to approximate those quantities. Note that based on the result of Proposition \ref{prop:invMGIG}, the importance sampling procedure for estimating mean of $\mathcal{MGIG}$, i.e., $\mathbb{E}_{\Lambda \sim\mathcal{MGIG}}[\Lambda]$, can also be applied to infer the reciprocal mean i.e. $\mathbb{E}_{\Lambda \sim\mathcal{MGIG}}[\Lambda^{-1}]$.

\begin{figure}
\centering
\subfigure[]{\includegraphics[width = 0.32\textwidth]{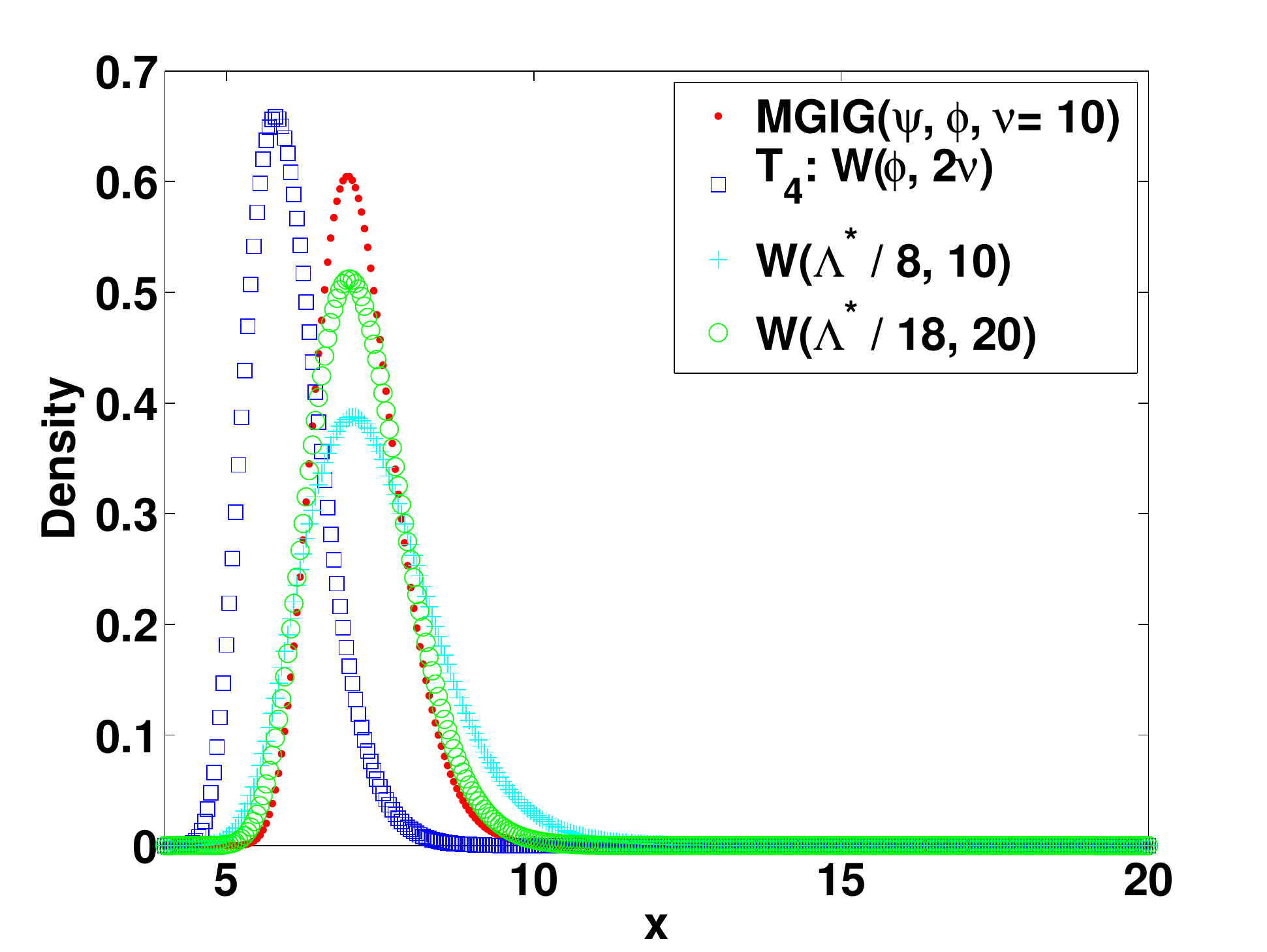} \label{fig:MGIGdens1dW}}
\hspace{-0.3cm}
\subfigure[]{\includegraphics[width = 0.32\textwidth]{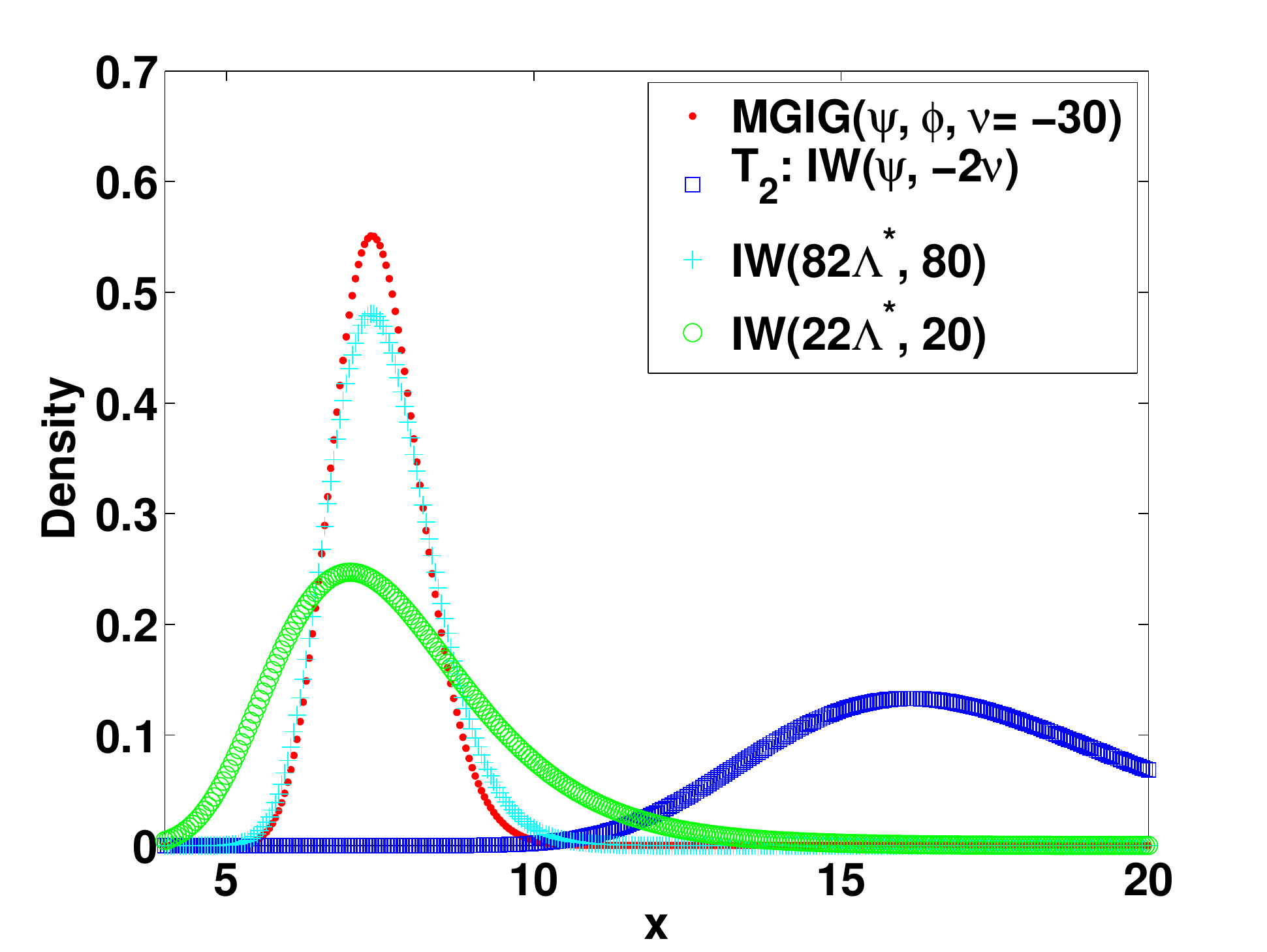} \label{fig:MGIGdens1dIW}}
\hspace{-0.3cm}
\subfigure[]{\includegraphics[width = 0.32\textwidth]{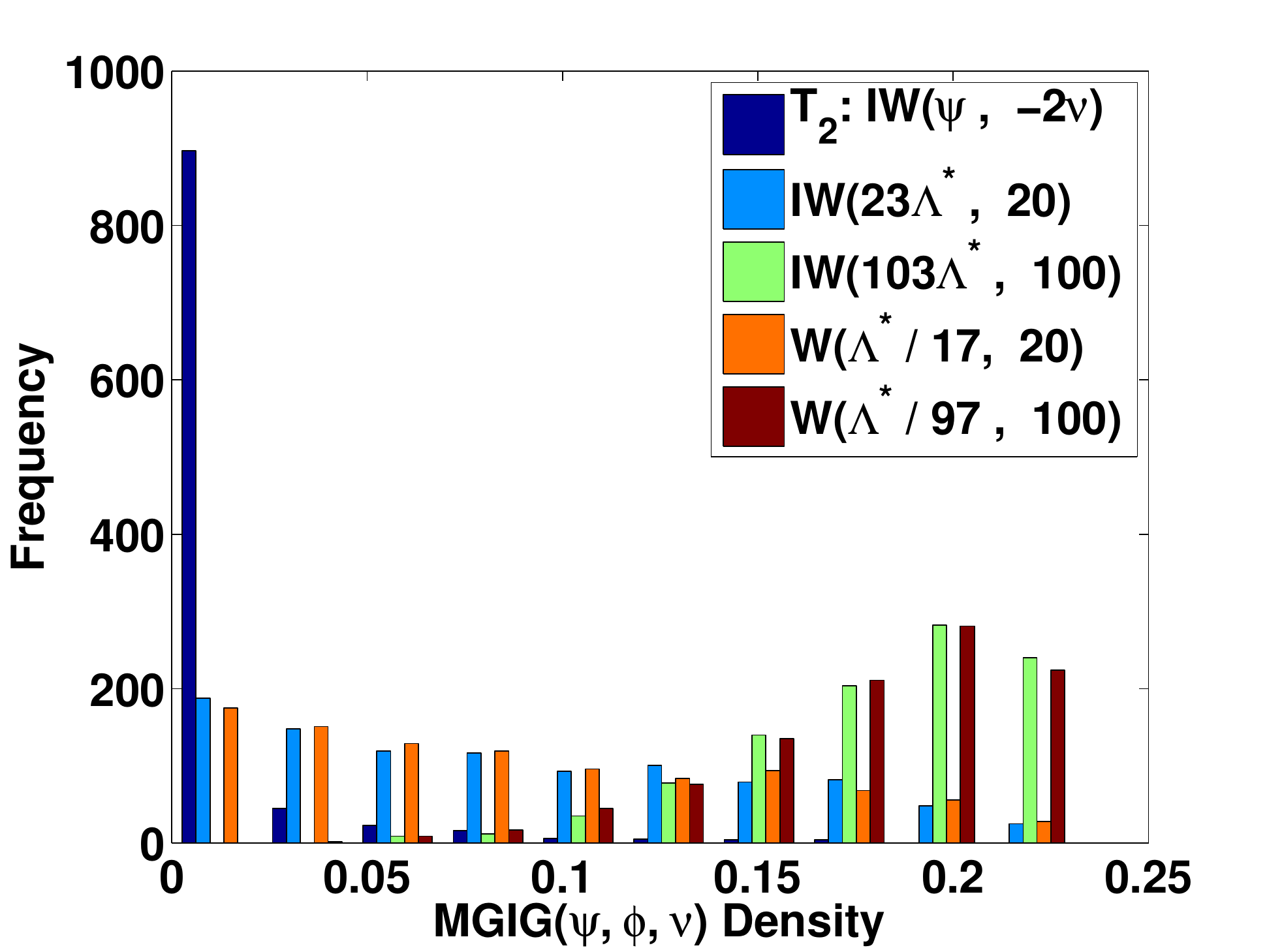} \label{fig:numSamWIW}}
\caption{(a,b) Comparison of different proposal distribution (a) Wishart ($\mathcal{W}$) and (b) Inverse Wishart ($\mathcal{IW}$) for sampling mean of $\mathcal{MGIG}_1(\Psi, \Phi, \nu)$ where $\Lambda^*$ is the mode of $MGIG$. The blue curves are the proposal distribution defined in \cite{yalz13,yoshii13} which can not recover the mode of the $\mathcal{MGIG}$ distribution. (c) Density of $\mathcal{MGIG}_2(\Psi, \Phi, \nu)$ for 1000 samples generated by each proposal distribution is calculated. More than $90\%$ of  samples generated by the previous proposal distribution in \cite{yalz13,yoshii13} ($\mathcal{IW}(\psi, -2\nu)$) have zero $\mathcal{MGIG}$ density leading to $ESS=40$. Whereas, the new proposal distribution $IW(23\Lambda^*, 20)$ has the $ESS=550$ which has a very similar shape to the target $\mathcal{MGIG}$ distribution.
}
\label{fig:MGIGvsWIW}
\end{figure}

An importance sampling procedure proposed in~\cite{yalz13,yoshii13}, where the $\mathcal{MGIG}$ is viewed as a product of Inverse Wishart and Wishart distributions and one of the multiplicands is used as the natural choice of the proposal distribution.
%which is not a suitable proposal distribution as illustrated in Figures~\ref{fig:MGIGvsWIW} and S1. 
%given the definition \ref{def:MGIGdens} of the $\mathcal{MGIG}$ distribution in (\ref{eq:MGIGdens}),
In particular, in \cite{yalz13,yoshii13}, the $\mathcal{MGIG}$ is viewed as
% considered $\mathcal{MGIG}_N$ density as product of two terms  as
\begin{align}
\mathcal{MGIG}_N(\Lambda | \Psi, \Psi, \nu) &\propto  \underbrace{e^{\Tr( -\frac{1}{2} \Phi \Lambda)}}_{T_1} \underbrace{\mathcal{IW}_N(\Lambda \given \Psi, -2\nu_u)}_{T_2} \label{eq:MGIGIW}\\
 &\propto  \underbrace{e^{\Tr( -\frac{1}{2} \Psi \Lambda^{-1})}}_{T_3} \underbrace{\mathcal{W}_N (\Lambda \given \Phi, 2\nu_u)}_{T_4}. \label{eq:MGIGW}
\end{align}
%where $T_1 = e^{Tr( \frac{-1}{2} \Psi \Lambda)} $, $T_2 = \mathcal{IW}_N(\Lambda \given \Psi, -2\nu_u)$, $T_3=e^{Tr( \frac{-1}{2} \Psi \Lambda^{-1})}$, and $T_4 =\mathcal{W}_N (\Lambda \given \Psi, 2\nu_u)$.
Note that $T_2$ is the Inverse Wishart distribution %$\mathcal{IW}_N(\Lambda \given \Psi, -2\nu_u)$
and $T_4 $ is the Wishart distribution,
%$\mathcal{W}_N (\Lambda \given \Psi, 2\nu_u)$
and there are efficient samplers for both of Wishart and Inverse Wishart distributions \cite{smho72}.
In~\cite{yalz13,yoshii13}, authors advocate using $T_2$ (or $T_4$) as the proposal distribution which simplify the weight calculation to the evaluation of $T_1$ (or $T_3$).
However, it is not studied how close $T_2$ (or $T_4$) are to the $\mathcal{MGIG}$ distribution in shape. For example, consider the $1-$dimensional $\mathcal{MGIG}$ distribution
\begin{align}
\mathcal{MGIG}_1(\Lambda \given 35, 10, 10) \propto \underbrace{e^{\Tr( -\frac{35}{2} \Lambda^{-1})}}_{T_3} \underbrace{\mathcal{W}_1 (\Lambda \given 10, 20)}_{T_4}.
\end{align}
In \cite{yalz13,yoshii13}, $T_4: \mathcal{W}_1 (\Lambda \given 10, 20)$ is considered as the proposal distribution, but the mode of $T_4$ is far away from the mode of $\mathcal{MGIG}_1(\Lambda \given 35, 10, 10)$ (Figure \ref{fig:MGIGvsWIW}(a)). As a result, samples drawn from $T_4$ will be on the tail of the $\mathcal{MGIG}_1(\Lambda \given 10, 20)$ distribution, and will end up getting low weights (importance) from the $\mathcal{MGIG}_1(\Lambda \given 10, 20)$ distribution. Such a sampling procedure will be wasteful, i.e., drawing samples from the tails of the target $\mathcal{MGIG}_1$  distribution, leading to a very low $ESS$. Similar behavior is observed with several different choices of parameters for the $\mathcal{MGIG}$, here we only show three of them in Figure \ref{fig:MGIGvsWIW} due to the lack of space.
% with other  with $\mathcal{MGIG}_1(\Lambda \given \Psi, \Phi, -30) \propto  \underbrace{e^{\Tr( -\frac{1}{2} \Phi \Lambda)}}_{T_1} \underbrace{\mathcal{IW}_1(\Lambda \given \Psi, -60)}_{T_2}$ where the mode of $T_2$ is far a way from the mode of $\mathcal{MGIG}_1(\Lambda \given \Psi, \Phi, -30)$ (Figure \ref{fig:MGIGvsWIW}(b)).
%if both of the parameters $\Phi$ and $\Psi$ are strong, e.g., with large spectral norms, $T_2$ (or $T_4$) is not a good proposal distributions for the $\mathcal{MGIG}_N$, since the mode of $T_2$ (or $T_4$) is going to be far away from the mode of $\mathcal{MGIG}_N$. As a result, samples drawn from $T_2$ (or $T_4$) will be on the tail of the $\mathcal{MGIG}_N$ distribution, and will end up getting low weights (importance) from the $\mathcal{MGIG}_N$ distribution (Figures \ref{fig:MGIGvsWIW}(a-b), and S1 in the appendix). Thus, the sampling procedure will be wasteful, i.e., drawing samples from the tails of the target  $\mathcal{MGIG}_N$  distribution, leading to a very low $ESS$ (Figure \ref{fig:numSamWIW}). Figure 

\subsection{Algebraic Riccati Equation}
An algebraic Riccati equation (ARE) is 
\begin{align}
A^TX + XA + X R X + Q = 0,
\label{eq:ARE}
\end{align}
where $A \in \mathbb{R}^{N \times N}, Q \in \mathbb{S}_{+}^N$, and $R \in \mathbb{S}_{+}^N$.
We associate a $2N \times 2N$ matrix called the Hamiltonian matrix $H$ with the ARE \eqref{eq:ARE} as
%\begin{eqnarray}
$H = \begin{bmatrix}
A  & R \\
-Q & -A^T
\end{bmatrix}$.
%\label{eq:AREhamMat}
%\end{eqnarray}
The Hamiltonian matrix has some useful properties appears in various control and filtering problems of continuous time systems. In particular, the ARE \eqref{eq:ARE} has a unique positive definite solution if and only if the associated Hamiltonian matrix $H$ has no imaginary eigenvalues (Section 5.6.3 of \cite{boba91}).

There have been offered various numerical methods to solve the ARE which can be reviewed in \cite{anmo07}. 
%schur \cite{arla84, door81, laub79}.
The key of numerical technique to solve ARE \eqref{eq:ARE} is to convert the problem to a stable invariant subspace problem of the Hamiltonian matrix i.e., finding the invariant subspace corresponding to the eigenvalues of $H$ with negative real parts.
In particular, consider $V = \begin{bmatrix}
X_1 \\
X_2
\end{bmatrix}$ to be a $H-$invariant subspace, i.e., $H V = V \Lambda$.
%We observe that $X$ satisfies \eqref{eq:ARE} if and only if 
%\begin{align}
%H
%\begin{bmatrix}
%X_1 \\
%X_2
%\end{bmatrix}  = 
%\begin{bmatrix}
%A  & R \\
%-Q & -A^T
%\end{bmatrix} 
%\begin{bmatrix}
%X_1 \\
%X_2
%\end{bmatrix} = 
%\begin{bmatrix}
%X_1 \\
%X_2
%\end{bmatrix} \Lambda.
%\end{align}
Assume $X_1$ is invertible, we then post multiply by $X_1^{-1}$ to obtain
\begin{align}
H V X_1^{-1} = V \Lambda X_1^{-1} \Rightarrow 
%\begin{bmatrix}
%A  & R \\
%-Q & -A^T
%\end{bmatrix} 
%\begin{bmatrix}
%X_1 \\
%X_2
%\end{bmatrix} 
%X_1^{-1}
%=
\begin{bmatrix}
A  & R \\
-Q & -A^T
\end{bmatrix} 
\begin{bmatrix}
I \\
X
\end{bmatrix}
%= 
%\begin{bmatrix}
%X_1 \\
%X_2
%\end{bmatrix} \Lambda X_1^{-1}
=
\begin{bmatrix}
I \\
X
\end{bmatrix} X_1 \Lambda X_1^{-1},
\end{align}
where $X=X_2 X_1^{-1}$. Multiplying both side by $\begin{bmatrix} -X &  I \end{bmatrix}$, we have
\begin{align}
\begin{bmatrix} -X &  I \end{bmatrix}
\begin{bmatrix}
A  & R \\
-Q & -A^T
\end{bmatrix} 
\begin{bmatrix}
I \\
X
\end{bmatrix} =
\begin{bmatrix} -X &  I \end{bmatrix}
\begin{bmatrix}
I \\
X
\end{bmatrix} X_1 \Lambda X_1^{-1} = 0.
\end{align}
Simplifying the left hand side we get the ARE \eqref{eq:ARE} which implies that $X=X_2X_1^{-1}$ is the solution of \eqref{eq:ARE}.

%Several different algorithms has been proposed for solving ARE such as \cite{arwL84,byer87,laub79,van81} which can be applied to find the mode of $\mathcal{MGIG}$ distribution. 
The usual ARE solvers such as the Schur 
vector method \cite{laub79},  SR methods \cite{buns86}, the matrix sign function \cite{bai98,byer87} require in general $O(n^3)$ flops~\cite{lclw13}. 
%Faster algorithms can be applied to solve ARE depending on the structure of $Phi$ or $\Psi$ \cite{lclw13}.
For special cases, faster algorithms such as \cite{lclw13} can be applied which solves such an ARE with 20k dimensions in seconds. In this paper, we use Matlab ARE solver (\textit{care}) to find the solution of ARE.

\subsection{PMF, PPCA, and Bayesian PCA}
\label{sec:PMFPPCA}
Here, we giv a review of PMF \cite{samn07}, Probabilistic PCA (PPCA) \cite{tibi99}, and Bayesian PCA (BPCA) \cite{bish99a}, to illustrate the similarity and differences between the existing ideas and our approach. Table \ref{tab:sumAlg} provides a summary of the algorithms. A related discussion appears in \cite{laur09}. All these models focus on an (partially) observed data matrix $X \in \mathbb{R}^{N \times M}$. Given latent factors $U \in \mathbb{R}^{N \times D}$ and $V \in \mathbb{R}^{M \times D}$, the rows of $X$ are assumed to be generated according to $\mathbf{x}_{:m} = U \mathbf{v}_m^T + \epsilon$, where $\epsilon \in \mathbb{R}^N$. The different models vary depending on how they handle distributions or estimates of the latent factors $U, V$.
 Without loss of generality, for all the analysis through the paper, we are considering a fat matrix $X$ where $M > N$.

%\begin{table}
%\caption{Summary of the algorithms.}
%\label{tab:sumAlg}
%\begin{tabular}{l| l| l| l}
%Algorithm         & prior of $U$ & prior of $V$ & Inference \\
%         \hline
%PPCA \cite{tibi99} & $N(0, \sigma_u \mathbb{I})$ & - &  Point Est. of $V$ \\
%BPCA \cite{bish99a} & $N(0, \sigma_u \mathbb{I})$ & $N(0, \sigma_v \mathbb{I})$ & Laplace Approx.\\
%PMF \cite{samn07}  &  $N(0, \sigma_u \mathbb{I})$ & $N(0, \sigma_v \mathbb{I})$ & Point Est. \\
%BPMF \cite{samn08b} &  $N(0, \Sigma_u)$ & $N(0, \Sigma_v)$ & MCMC \\
%CMC-PMF & $N(0, \sigma_u \mathbb{I})$ & $N(0, \sigma_v \mathbb{I})$ & CMC \\
%\hline
%\end{tabular}
%\vspace{-0.5cm}
%\end{table}
%The sparse matrix $X \in \mathbb{R}^{N \times M}$ is approximated by a low-rank matrix $\hat{X} = U V^T$ where $U \in \mathbb{R}^{N \times D} $ and $V \in \mathbb{R}^{M \times D}$. %Each user $n$, and movie $m$ in $X$ are characterized by latent factors $u_n\in \mathbb{R}^D$ and $v_m \in \mathbb{R}^D$, respectively \cite{jans13, kobv09}.
%%Once the latent factors have been estimated, for each missing entry $x_{nm}$, the inner product of $u_n$ and $v_m$ gives the prediction for this entry.
%% The generative process of PMF is as follows:
%%\vspace{-0.2in}
%The likelihood is given as
%\begin{align}
% p \left( X| U, V \right)  ~ \prod_{n,m}  [ \mathcal{N} (x_{nm} \big | \langle\mathbf{u}_n, \mathbf{v}_m \rangle, \sigma^2 )]^{\delta_{nm}}
% \label{eq:PMFlik}
%\end{align}

\noindent \textbf{PMF and BPMF:} In PMF \cite{samn07}, one assumes independent Gaussian priors for all latent vectors $\mathbf{u}_n$ and $\mathbf{v}_m$, i.e.,  $\mathbf{u}_n \sim \cN (0, \sigma^2_u \mathbb{I}), [n]_1^N$ and $\mathbf{v}_{m} \sim \cN(0, \sigma^2_v \mathbb{I}), [m]_1^M$. Then, one obtains the following posterior over $(U,V)$
%\vspace{-0.1cm}
\begin{align}
%\label{eq: postUV}
 p \left( U, V | X ,\sigma^2, \sigma_u^2, \sigma_v^2	\right)  =
 \prod_{n,m}  [ \mathcal{N} (x_{nm} \big | \langle\mathbf{u}_n, \mathbf{v}_m \rangle, \sigma^2 )]^{\delta_{nm}}
  \prod_{n} \mathcal{N} ( \mathbf{u}_n \big | 0, \sigma_u^2 \mathbb{I})  \prod_{m} \mathcal{N} ( \mathbf{v}_{:m} \big | 0, \sigma_v^2 \mathbb{I})~,
%\nonumber
\end{align}
%\vspace{-0.4cm}
where $\delta_{nm}= 0$ if $x_{nm}$ is missing. PMF obtains point estimates
$(\hat{U},\hat{V})$ by maximizing the posterior (MAP), based on alternating optimization over $U$ and $V$~\cite{samn07}.

Bayesian PMF~(BPMF)~\cite{samn08b} considers independent Gaussian priors over latent factors with full covariance matrices,~i.e.,~$\mathbf{u}_n \sim \cN (0, \Sigma_u), [n]_1^N$~and~$\mathbf{v}_{m} \sim \cN(0, \Sigma_v), [m]_1^M$. Inference is done using Gibbs sampling to approximate the posterior $P(U, V | X)$. At each iteration, $U$ is sampled from the conditional probability of $p(U | V, X)$, followed by sampling $V$ from $p(V | U, X)$ using the updated matrix $U$ at the current iteration.

%The Gibbs sampling algorithm like other MCMC approaches asymptotically converges to true posterior distribution. However, they are computationally expensive especially on large data sets.

%Bayesian PCA is the Bayesian formulation of PCA by introducing prior distribution over the model parameters \cite{bish99a}.

%Consider a data set $X$ of observed $M$ dimensional vectors $X = \{\mathbf{x}_n\}_{n=1}^N$. The observed variable $\mathbf{x}_n$ is defined as a linear transformation of a latent variable $\mathbf{u_n} \in \mathbb{R}^{D}$ as $\mathbf{x}_n = \mathbf{u}_nV^T + \epsilon$ where $V \in \mathbb{R}^{M \times D}$ and $\epsilon \sim \mathcal{N}(0, \sigma^2I)$. Denote set of all latent variables $\mathbf{u}_n$ as $U = \{\mathbf{u}_n\}_{n=1}^N$.

\noindent {\bf Probabilistic PCA:} In PPCA \cite{tibi99}, one assumes independent Gaussian prior over $\u_n$, i.e., $\u_n \sim \cN(0,\sigma_u^2 \I)$, but $V$ is treated as a parameter to be estimated. In particular, in PPCA, $V$ is chosen so as to maximize the marginalized likelihood of $X$ given by
\begin{align}
p\left( X \given V \right)  = \int_U p(X| U, V) p(U) dU 
 = \prod_{n=1}^N \mathcal{N} (\mathbf{x}_n | 0, \sigma_u^2VV^T+\sigma^2 \mathbb{I}).  \label{eq:PPCA}
\end{align}
Interestingly, as shown in \cite{tibi99}, the estimate $\hat{V}$ can be obtained in closed form.
For such a fixed $\hat{V}$, the posterior distribution over $U|X,\hat{V}$ can be obtained as:
%\vspace{-0.3cm}
\begin{align}
p(U | X, \hat{V}) =  \frac{p(X|U, \hat{V}) p(U)}{p(X|\hat{V})} = \prod_{n=1}^N \mathcal{N} \left( \u_n | \Gamma^{-1} \hat{V}^T \mathbf{x}_n , \sigma^{-2} \Gamma \right), \label{eq:pcaPost}
\end{align}
where $\Gamma = \hat{V}^T\hat{V} +  \sigma_u^{-2} \sigma^{-2} \mathbb{I}$.
Note that the posterior of the latent factor $U$ in (\ref{eq:pcaPost}) depends on both  $X$ and $\hat{V}$.
For applications of PPCA in visualization, embedding, and data compression, any point $x_n$ in the data space can be summarized by its posterior mean $E[\mathbf{u}_n | \mathbf{x}_n, \hat{V}]$ and covariance $Cov(\mathbf{u}_n | \hat{V})$ in the latent space.

%In PPCA, the optimum value of $V$  is obtained by maximizing marginalized likelihood (\ref{eq:PPCA}) with respect to $V$.
%, where marginalizing $Z$ provides the Gaussian marginalized likelihood as
%In PPCA, the mapping matrix $W$ is considered as a model parameter where it optimum value denoted by $W_{ML}$ is obtained by maximizing (\ref{eq:PPCA}) with respect to $W$.
\noindent \textbf{Bayesian PCA:} In Bayesian PCA \cite{bish99a}, one assumes independent Gaussian priors for all latent vectors $\mathbf{u}_n$ and $\mathbf{v}_m$, i.e.,  $\mathbf{u}_n \sim \cN (0, \sigma^2_u \mathbb{I})$ and $\mathbf{v}_{m} \sim \cN(0, \sigma^2_v \mathbb{I})$, $[m]_1^M$. Bayesian posterior inference by Bayes rule considers $p(U,V|X) = p(X|U,V) p(U) p(V)/p(X)$, which includes the intractable partition function
\begin{equation}
p(X) = \int_U \int_V p(X|U,V) p(U) p(V) dU dV~.
\end{equation}
\noindent The literature has considered approximate inference methods, such as variational inference \cite{bish99b}, gradient descent optimization \cite{laur09}, MCMC \cite{samn08b}, or Laplace approximation \cite{bish99a, mink00}.
%have been applied to estimate the model parameters.
%Bishop introduced a Gaussian prior over the other model parameter  $V$, $p(V) \sim \mathcal{N}(0, \sigma_v^2 \mathbb{I})$ which provides a full Bayesian version of PCA. %Since marginalizing the above probability (\ref{eq:PPCA}) over both $Z$ and $W$ is intractable, Laplace approximation is used to estimate the parameters.
%In Bayesian PCA, marginalizing the likelihood $p(X| U, V)$ over both $U$ and $V$ is not tractable. %the latent matrices $U$ and $V$ are estimated by applying approximation approaches like %the maximum a posterior (MAP) approach.
%It has been showed that the unconstrained PMF is equivalent to Bayesian PCA \cite{laur09}.
%The equivalence between PMF and Bayesian PCA is shown by setting the latent matrix $U$, and mapping matrix $V$ of Bayesian PCA to the latent matrices $U$ and $V$ in PMF, respectively. Note that, in absence of missing values, the likelihood (\ref{eq:PMFlik}) can be written as
%\begin{align}
%p \left( X | U, V, \sigma^2 \right)  =  \prod_{n}  \mathcal{N} (x_{n} \big | \mathbf{u}_n V^T, \sigma^2 \mathbb{I})
%\end{align}
%which is similar to the model generator of Bayesian PCA.
%Optimizing (\ref{eq:PPCA}) with respect to $W$ is called Probabilistic PCA.

While PPCA and Bayesian PCA were originally considered in the context of embedding and dimensionality reduction, PMF and BPMF have been widely used in the context of matrix completion where the observed matrix $X$ has many missing entries. Nevertheless, as seen from the above exposition, the structure of the models are closely related (also see \cite{laur09,lawr05}).

%In this work, we show that one of the latent matrices in PMF can be marginalized or `collapsed' yielding a Matrix Generalized Inverse Gaussian ($\mathcal{MGIG}$) \cite{babj82, butl98} posterior distribution over the other matrix i.e., $p(V|X)$ in PMF or $p(W|X)$ in Bayesian PCA denoting as the marginalized posterior distribution.
%The characterization leads to an efficient Collapsed Monte Carlo (CMC, not MCMC) inference algorithm for such models. The inference is done by marginalizing one of the latent matrices and doing a direct Monte Carlo sampling from the posterior of the other latent matrix, by using some recent advances in sampling from the $\mathcal{MGIG}$ distribution \cite{yalz13}.

\begin{table}[t]
\begin{center}
\caption{Summary of low rank matrix factorization algorithms illustrated as the handled distribution where (.) is the inference procedure. CF denotes Closed Form.
MCMC alternately sample both $U$ and $V$ from a Markov chain with the joint posterior $p(U,V|X)$. CMC only samples the smaller $U$ matrix directly from $p(U|X)$.
}
\label{tab:sumAlg}
\begin{tabular}{l l l}
{\textbf {Algorithm}}        & {\textbf {Inference of $U$}} &{\textbf { Inference of $V$}}  \\
         \hline
{\textbf {PPCA}} \cite{tibi99} & {\small{$p(U|X, \hat{V})$~(CF)}}&{\small{Point Est.~ $\hat{V}$ (ML)}} \\
{\textbf {BPCA}} \cite{bish99a} & {\small{$p(U,V|X)$~(Approx.)}}&{\small{$p(U,V|X)$~(Approx.)}} \\
{\textbf {PMF}} \cite{samn07}  &  {\small{Point Est.~(MAP)}}&{\small{Point Est.~(MAP)}} \\
{\textbf {BPMF}} \cite{samn08b} & {\small{$p(U,V|X)$~(MCMC)}}&{\small{$p(U,V|X)$~(MCMC)}} \\
{\textbf {CMC-PMF (ours)}} & {\small{$p(U|X)$~(CMC)}}&{\small{$p(V|X)$~(CMC)}} \\
\hline
\end{tabular}
\end{center}
\end{table}

\section{$\boldsymbol{\mathcal{MGIG}}$ Properties and Sampling}
\label{sec:mgig}
%%The $\mathcal{MGIG}$ distribution has been recently used as a prior for covariance matrix~\cite{} because it can be considered as mixture of wishart and inverse wishart distribution. 
%%we show that $\mathcal{MGIG}$ is uni-modal and provide an efficient sampler using Monte Carlo estimation.
%We first give a brief background on $\mathcal{MGIG}$ distribution and importance sampling technique. 
%%Then, we show that the $\mathcal{MGIG}$ distribution is unimodal where the  mode can be obtained by solving an {\em Algebraic Riccati Equation (ARE)}~\cite{boba91}. 
%Then, we introduce a new importance sampler to infer the mean of $\mathcal{MGIG}$.
%%using the mode matching idea. 
%%Next, we give a brief overview of importance sampling method, then we provide the 
%%Our proposed distribution is an inverse Wishart (or Wishart) distribution with the same mode as $\mathcal{MGIG}$ distribution. As  illustrated in Figures \ref{fig:MGIGvsWIW} and S1, the new proposal distribution
%%requires less number of samples than the sampler in \cite{yalz13} for the mean estimation.
%
%
%\subsection{$\boldsymbol{\mathcal{MGIG}}$ Distribution is Unimodal}
Some properties of the $\mathcal{MGIG}$ distribution and its connection with Wishart distribution has been studied in~\cite{butl98,seshadri03,seshadri08}.~However, to best of our knowledge, it is not yet known if the distribution is unimodal and how to obtain the mode of $\mathcal{MGIG}$. In the following Lemma we show that the $\mathcal{MGIG}$ distribution is unimodal.

%\subsection{$\boldsymbol{\mathcal{MGIG}}$ is Unimodal}
\begin{Lemma}
Consider the $\mathcal{MGIG}$ distribution $\mathcal{MGIG}_N(\Lambda | \Psi, \Phi, \nu)$. 
The mode of  $\mathcal{MGIG}$ distribution is the solution of the following Algebraic Riccati Equation (ARE)
%Then, the $\mathcal{MGIG}$ distribution is unimodal and the mode can be found by solving the following Algebraic Riccati Equation (ARE)
\begin{align}
-2\alpha \Lambda + \Lambda \Phi \Lambda -  \Psi = 0,
\label{eq:AREmodeMGIG}
\end{align}
where $\alpha = (\nu - \frac{N+1}{2})$. ARE in \eqref{eq:AREmodeMGIG} has a unique positive definite solution, thus the $\mathcal{MGIG}$ distribution is a unimodal distribution.
\end{Lemma}
\proof
The $\log$-density of $\mathcal{MGIG}_N(\Lambda | \Psi, \Phi, \nu)$  is 
%propotional to
\begin{align}
%\hspace{-0.2cm}
%& \log \mathcal{MGIG} \\
\log f(\Lambda) = \alpha \log |\Lambda | - \frac{1}{2} \Tr(\Psi \Lambda^{-1} + \Phi \Lambda) + C,
\label{eq:logLikMGIG}
\end{align}
%Set derivative of (\ref{eq:logLikMGIG}) to zero, to find the mode of  $\mathcal{MGIG}$.
where $\alpha = (\nu - \frac{N+1}{2})$, and $C$ is a constant which does not depend on $\Lambda$. The mode of $\mathcal{MGIG}_N$ is obtained by setting derivative of (\ref{eq:logLikMGIG}) to zero. The derivative is a quadratic matrix equation as follows
\begin{align}
\nabla f(\Lambda) = -2\alpha \Lambda + \Lambda \Phi \Lambda -  \Psi = 0.
%-(\nu - \frac{N+1}{2}) \Lambda + \frac{1}{2} \Lambda \Phi \Lambda - \frac{1}{2} \Psi = 0.
\label{eq:modMGIG}
\end{align}
Note that \eqref{eq:modMGIG} is a special case of ARE \eqref{eq:ARE}.
%written as $A^TX + XA + \gamma^{-1} X BB^T X + \gamma^{-1} C^TC = 0 $. 
%\begin{eqnarray}
%H_\gamma = \begin{pmatrix}
%A  & \gamma^{-1}BB^T \\
%-\gamma^{-1} C^TC & -A^T
%\end{pmatrix}.
%\label{eq:AREhamMat}
%\end{eqnarray}
The associated Hamiltonian matrix for (\ref{eq:modMGIG}) is
%\begin{eqnarray}
${H} = \begin{bmatrix}
- \alpha \mathbb{I}_N  & \Phi \\
\Psi & \alpha \mathbb{I}_N
\end{bmatrix}$.
%\label{eq:MGIGhamMat}
%\end{eqnarray}
Recall that ARE has a unique positive definite solution if and only if the associated Hamiltonian matrix $H$ has no imaginary eigenvalues (Section 5.6.3 of \cite{boba91}). Thus, to show the unimodality of $\mathcal{MGIG}$, it is enough to show that
%(\ref{eq:MGIGhamMat}) has no imaginary eigenvalue or equivalently
the corresponding characteristic polynomial $| {H} - \lambda \mathbb{I}_{2N}  | = 0$ has no imaginary solution.
\begin{eqnarray}
| {H} - \lambda \mathbb{I}_{2N} | &=&
\left|
\begin{matrix}
-(\alpha + \lambda)\mathbb{I}_N  & \Phi \\
\Psi & (\alpha - \lambda) \mathbb{I}_N
\end{matrix} \right|
\notag \\
&=& \left| -(\alpha + \lambda)\mathbb{I}_N \right| \left|  (\alpha - \lambda) \mathbb{I}_N + (\alpha + \lambda)^{-1} \Psi \Phi \right|  \notag \\
&=&\left| (\alpha - \lambda) \mathbb{I}_N \right| \left|  -(\alpha + \lambda)\mathbb{I}_N  -  (\alpha - \lambda)^{-1}\Phi \Psi \right|  \notag \\
&=&\prod_{i=1}^N  \{- (\alpha^2 - \lambda^2) - \tilde{\lambda}_i\} = 0,
% = \left| -(\alpha^2 - \lambda^2) \left( \mathbb{I}_M  + \frac{\Phi \Psi}{(\alpha^2 - \lambda^2) } \right) \right|
%  \notag \\
% & = (\alpha^2 - \lambda^2) ^{M} \left| \mathbb{I}  + \frac{\Phi \Psi}{(\alpha^2 - \lambda^2) }  \right| = 0
\label{eq:detMGIG}
\end{eqnarray}

%Given that $\lambda \neq \pm \alpha$, t
%The solution of (\ref{eq:detMGIG}) is  solution of
%$\prod_{i=1}^M (1+\frac{\tilde{\lambda}_i}{\alpha^2 - \lambda^2}) = 0$
\noindent which yields to $\lambda^2 = \tilde{\lambda}_i + \alpha^2$ where $\tilde{\lambda}_i$ is the $i^{th}$ eigenvalue of $\Phi \Psi$. %Note that $\lambda \neq \pm \alpha$ due to the inversion in (\ref{eq:detMGIG}).
Note $\tilde{\lambda}_i > 0$ since $\Phi$ and $\Psi$ are positive definite and product of two positive definite matrix has positive eigenvalue . As a result, (\ref{eq:detMGIG}) has no imaginary solution and ${H}$ does not have any imaginary eigenvalue. 
As a result, ARE in \eqref{eq:modMGIG} has a unique positive definite solution. This completes the proof.
\qed

%\subsection{Importance Sampling for $\boldsymbol{\mathcal{MGIG}}$}
%\label{sec:sampleMeanMGIG}
\noindent {\bf{Importance Sampling for $\boldsymbol{\mathcal{MGIG}}$: }}Since $\mathcal{MGIG}$ is a unimodal distribution, we propose an efficient importance sampling procedure for $\mathcal{MGIG}$ by mode matching.
%A good choice of the proposal distribution is the one with the same mode as the target distribution (mode matching) where the proposal $q(x)$ is large in a region where $p(x)$ is large providing a good estimate of the expectation.
We select a proposal distribution over space of positive definite matrices by matching the proposal's mode to the mode of $\mathcal{MGIG}$ (mode matching) which aligns the proposal and $\mathcal{MGIG}$ shapes. Mode matching is a good choice of the proposal as the proposal $q(x)$ is large in a region where the target distribution $\mathcal{MGIG}$ is large leading to a good estimate of the expectations $\mathbb{E}_{\Lambda \sim\mathcal{MGIG}}[\Lambda]$ or $\mathbb{E}_{\Lambda \sim\mathcal{MGIG}}[\Lambda^{-1}]$.
%using mode matching which requires to first obtain the mode of $\mathcal{MGIG}$.
%We propose to choose the proposal distribution with domain of symmetric positive definite matrices. We match the shape of proposal distribution with $\mathcal{MGIG}$ by choosing the mode of proposal distribution the same as $\mathcal{MGIG}$ distribution.
An example of such proposal distribution is Inverse Wishart or Wishart distribution.

Let $\Lambda^*$ be the mode of $\mathcal{MGIG}_N(\Lambda | \Psi, \Phi, \nu)$ which can be found by solving the ARE \eqref{eq:modMGIG}. The mode of  Inverse Wishart $\mathcal{W}_N(\Lambda | \Sigma, \rho)$ distribution is $\Sigma^* = (\rho -N -1)\Sigma$. To match the mode of $\mathcal{W}_N(\Lambda | \Sigma, \rho)$ with that of $\mathcal{MGIG}_N(\Lambda | \Psi, \Phi, \nu)$, we choose the scale parameter $\Sigma$ of the Wishart distribution by setting $\Sigma^* = \Lambda^*$. In particular,
\begin{align}
\Sigma^* = \Lambda^* = (\rho -N -1)\Sigma \quad \Rightarrow \quad \Sigma = \frac{\Lambda^*}{\rho -N -1}.
\end{align}
Thus, we suggest using $\mathcal{W}_N(\frac{\Lambda^*}{\rho-N-1}, \rho)$ as the proposal distribution. At each iteration, we draw a sample $\Lambda_i \sim \mathcal{W}_N(\frac{\Lambda^*}{\rho-N-1}, \rho)$, and calculate the importance weight as $w(\Lambda_i) =\frac{\mathcal{MGIG}_N(\Lambda_i | \Psi, \Phi, \nu)}{\mathcal{W}_N(\Lambda_i | \Sigma, \rho)}$. More specifically, the density of Wishart distribution is
\begin{align}
q(\Lambda) = \frac{q^*(\Lambda)}{Z_q}, \qquad \text{where} \quad
q^*(\Lambda) = |\Lambda|^{\frac{\rho-N-1}{2}}\exp \{-\frac{1}{2}\Tr(\Sigma^{-1} \Lambda) \}.
\end{align}
Then, the importance weight can be calculated as
\begin{align}
w(\Lambda_i) = & \frac{\mid \Lambda_i \mid ^{ \nu - (N+1)/2} \exp \{-\frac{1}{2} \Tr( \Psi \Lambda_i^{-1} + \Phi \Lambda_i ) \}}{ |\Lambda_i|^{\frac{\rho-N-1}{2}}\exp \{-\frac{1}{2}\Tr(\Sigma^{-1} \Lambda_i) \}} \\
 = & \mid \Lambda_i \mid ^{ \nu - \frac{\rho}{2}}  \exp \left \{-\frac{1}{2} \Tr \left(\Psi \Lambda_i^{-1} + [\Phi -\Sigma^{-1}] \Lambda_i \right)\right \}.
\end{align}
As a result, we can approximate the sample mean as
\begin{align}
\mathbb{E}_{\Lambda \sim\mathcal{MGIG}}[f(\Lambda)] = \frac{\sum_{i=1}^S w(\Lambda_i) f(\Lambda_i)}{\sum_{j=1}^S w(\Lambda_j)}.
\end{align}
%Mixture importance sampling can be provided by considering multiple $\mathcal{IW}$ and $\mathcal{W}$ with different $\rho$,  and report weighted average over all drawn samples as the mean.
Note that the weight calculation requires to calculate the inverse and determinant of sampled matrix $\Lambda_i$. However, as illustrated in Algorithm \ref{alg:wishRnd}, the random samples generator from $\mathcal{W}$ \cite{smho72} returns the upper triangular matrix $R$ where $\Lambda = R^TR$. Hence the inverse and determinant of $\Lambda$ can be calculated efficiently from the inverse and diagonal of the triangular matrix $R$, respectively.
Therefore, the cost of weight calculation is reduced to the cost of solving a linear system and upper triangular matrix production at each iteration.

A similar argument holds when the proposal distribution is an Inverse Wishart distribution. In particular, the mode of Inverse Wishart $\mathcal{IW}_N(\Sigma, \rho)$ distribution is $\frac{\Sigma}{\rho+N+1}$. Thus $\mathcal{IW}_N(\rho+N+1) \Lambda^*, \rho)$ is another suitable choice of the proposal distribution.

Figure \ref{fig:MGIGvsWIW} illustrates that the proposed importance sampling outperforms the one in \cite{yalz13,yoshii13} for three examples of $\mathcal{MGIG}$. In particular, more than $90\%$ of samples drawn from the proposal distribution $T_2$ in \cite{yalz13,yoshii13} have zero weights leading to $ESS=40$ (Figure \ref{fig:MGIGvsWIW} (c)). Whereas, our proposal distribution achieved $ESS=550$ leading to a better approximation of the mean of $\mathcal{MGIG}$. Similar behavior is observed with several different choices of parameters for the $\mathcal{MGIG}$.

 \begin{center}
\begin{algorithm}[t]
\caption{Random Generator of $\mathcal{W}_N(\Lambda | \Sigma, \rho, L)$ \cite{smho72}}
\label{alg:wishRnd}
\begin{algorithmic}[1]
  \algnotext{EndFor}
  \State Note $L^TL=\Sigma$ is the Cholesky factorization of $\Sigma$
  \State $P_{ii} \sim \sqrt{ \chi^2 (\rho - (i-1))}$ for all $i = 1 \cdots N$.
  \State $P_{ij} \sim \mathcal{N}(0,1)$ for $i < j$.  \Comment{$P$: upper triangular}
  \State $R = P  L$.
  \State Return $\Lambda = R^T R$.
  \State Return $\Lambda^{-1} = {R^{-1}} {R^{-1}}^T $.
  \State Return $|\Lambda| = [\prod_{i=1}^N R_{ii} ]^2$.
\end{algorithmic}
\end{algorithm}
\end{center}

\section{Connection of $\boldsymbol{\mathcal{MGIG}}$ and Bayesian PCA}
% An application of $\boldsymbol{\mathcal{MGIG}}$
\label{sec:application}
In this section, we illustrate %that one of the latent matrices in PMF can be marginalized or ‘collapsed’ yielding a $\mathcal{MGIG}$ posterior distribution over the other latent matrix.
that the mapping matrix $V$ in Bayesian PCA can be marginalized or `collapsed' yielding a Matrix Generalized Inverse Gaussian ($\mathcal{MGIG}$) \cite{babj82, butl98} posterior distribution over the latent matrix $U$ denoting as the marginalized posterior distribution. Then, we explain the derivation of the marginalized posterior for data with missing values, followed by a collapsed Monte Carlo Inference for PMF.
%Here, we show an application of $\mathcal{MGIG}$ distribution in latent factor models. In particular, we show the marginalized posterior in PMF (BPCA) has a $\mathcal{MGIG}$ distribution.

%\subsection{PMF, PPCA, and Bayesian PCA}
%\label{sec:PMFPPCA}
%\input{bayesPCA}
\subsection{Closed form Posterior Distribution in Bayesian PCA}
\label{sec:colPMF}
The key challenge in models such as Bayesian PCA or Bayesian PMF is that joint marginalization over both latent factors $U, V$ is intractable. Probabilistic PCA gets around the problem by considering one of the variables, say $V$, to be a constant. In this section, we show that one can marginalize or `collapse' one of the latent factors, say $V$, and obtain the marginalized posterior $P(U|X)$ over the other variable denoted.
In fact, we obtain the posterior with respect to the covariance structure $\Lambda_u = \beta_u \I + UU^T$, for a suitable constant $\beta_u$, which is sufficient to do Bayesian inference on new test points $x_{\text{test}}$.

We start with an outline of the analysis. Note that
\beq
p(U|X) \propto p(U)P(X|U) = p(U) \int_V P(X|U,V) p(V) dV~,
\label{eq:upost1}
\eeq
%\vspace{-0.5cm}
\noindent and, based on the posterior over $U$, one can obtain the probability on a new point as
%\vspace{-0.5cm}
\beq
p(x_{\text{test}}|X) = \int_U p(x_{\text{test}}|U) p(U|X) dU~.
\label{eq:upred1}
\eeq
%\vspace{-0.7cm}
\noindent Next, we show that the posterior over $U$ as in \myref{eq:upost1}, rather the distribution over $\Lambda_u = \beta_u \I + UU^T$, can be derived analytically in {\em closed form}. The distribution is the Matrix Generalized Inverse Gaussian ($\mathcal{MGIG}$) distribution.
Now, similar to (\ref{eq:PPCA}), marginalizing $V$ gives %the marginalized likelihood as
\begin{align*}
\hspace{-2cm} & p\left( X \given U \right)  = \int_V p(X| U, V) p(V) dV    \label{eq:margW} \\
& =    \prod_{m=1}^M \int_{\mathbf{v}_m } \mathcal{N} \left( {\mathbf{x}}_{:m} \given U \mathbf{v}_{m}^T  , \sigma^2 \mathbb{I} \right) \mathcal{N} \left(\mathbf{v}_m \given 0, \sigma^2_v \mathbb{I} \right) \ud \mathbf{v}_m   \\
%& = \prod_{m=1}^M  \mathcal{N}\left( \mathbf{x}_{:m} \given 0,  \sigma^2 \mathbb{I} + \sigma^2_v {U} {U}^T \right)
& = \prod_{m=1}^M \mathcal{N}\left( \mathbf{x}_{:m} \given 0,  \sigma_v^2 \Lambda_u \right),
\end{align*}
where $\Lambda_u = \beta_v \mathbb{I} + UU^T $ and $\beta_v = \frac{\sigma^2}{\sigma^2_v}$.
Then, the marginalized posterior of $U$ is calculated as
\begin{eqnarray}
p(U  \given   X ) \propto p({X} | U) ~ p(U) 
 &\propto & {\mid \Lambda_u \mid}^{-M/2}  \exp \left\{ -\frac{1}{2\sigma^2_v}  { \sum_{m=1}^M \mathbf{{x}}_{:m}^T ~\Lambda_u^{-1} ~\mathbf{x}_{:m}} \right\} \nonumber \\
&& \quad \times \exp \left\{  -\frac{1}{2 \sigma_u^2} \Tr(UU^T + \beta_u \mathbb{I} - \beta_u \mathbb{I}) \right\} \notag \\
 & =& {\mid \Lambda_u \mid}^{-M/2}  \exp \left\{ \frac{-\Tr\left(\Lambda_u^{-1} \sum_{m=1}^M \mathbf{x}_{:m} \mathbf{x}_{:m}^T\right)}{2\sigma_v^2}
\right\}   \notag \\
 \label{eq:factLam}
 && \quad \times \exp \left\{ \frac{-\Tr (\Lambda_u )}{2 \sigma_u^2}  \right\} \times \exp \left\{\frac{\Tr(\beta_u \mathbb{I})}{2\sigma_u^2}  \right\}  \\
 &=& ~{\mid \Lambda_u \mid}^{-M/2}  \exp \left\{ \Tr( -\frac{1}{2} \Lambda_u^{-1} \Psi_u -\frac{1}{2} \Lambda_u \Phi_u ) \right\}  \nonumber \\
&\sim &  \mathcal{MGIG} ( \Lambda_u \given \Psi_u, \Phi_u, \nu_u)~, \label{eq: margPostV}
\end{eqnarray}
where $\Psi_u = \frac{1}{\sigma_v^2} {X} {X}^T$, $\Phi_u = \frac{1}{\sigma_u^2} \mathbb{I}$, and $\nu_u = \frac{N-M+1}{2}$.

%Note that in case of no missing value, PMF is equivalent to Bayesian PCA \cite{bish99a}.
Therefore, by marginalizing or collapsing $V$, the posterior over $\Lambda_u = \beta_v \mathbb{I} + UU^T $
corresponding to the latent matrix $U$ can be characterized exactly with a $\mathcal{MGIG}$ distribution with parameters depending only on $X$. Note that this is in sharp contrast with
(\ref{eq:pcaPost}) for PPCA, where the posterior covariance of $\mathbf{u}_n$ is $\sigma^{-2} \Gamma$ which in turn depends on the point estimate for $\hat{V}$.

%With the same argument, one can marginalize $U$ to obtain the marginalized posterior of $V$ as
%\begin{align}
% p(X | V) &=& \sim \prod_{n=1}^N ~  \mathcal{N}\left( \mathbf{x}_{n} \given 0, ~ \Lambda_v \right)
% \label{eq:margLikV} \\
%p(\Lambda_v | X ) &\sim& \mathcal{MGIG} ( \Lambda_v \given \psi_v, \phi_v, \nu_v)
%\label{eq:PostlamV}
%\end{align}
%where  $\Lambda_v = \beta_v \mathbb{I} + VV^T $ and $\beta_v = \frac{\sigma^2}{\sigma^2_u}$,
%$\psi_v = \frac{1}{\sigma_u^2} {X}^T {X}$, $\phi_v = \frac{1}{\sigma_v^2} \mathbb{I}$, and $\nu_v = \frac{M-N+1}{2}$.

\subsection{Posterior Distribution with Missing Data}
\label{sec:missPostPMF}
%Assume some values of $X$ are missing
In this section, we consider the matrix completion setting, when the observed matrix $X$ has missing values.
%First,
%provide an estimator as well as proof of con- vergence for the missing data.
%For the matrix completion setting, when the observed matrix has missing values, the posterior distribution can not be directly obtained as MG I G in closed form. We propose zero-padding the data matrix to get the MG I G form, and prove that the covariance of the zero-padded matrix approaches that of the true covariance in spectral norm sense as samples increase.
In presence of missing values, the likelihood of the observed sub-vector in any column of $X$ is given as
\begin{align}
p \left( \mathbf{x}_{n_m, m} \given U, V \right) &= \mathcal{N} \left( \mathbf{x}_{n_m, m} \given \tilde{U}_m \mathbf{v}_m^T  , \sigma^2 \mathbb{I} \right).
\end{align}
where $n_m$ is a vector of size $\tilde{N}_m$ containing indices of non-missing entries in column $m$ of $X$,  and $\tilde{U}_m$ is a sub-matrix of $U$ with size of $\tilde{N}_m \times D$ where each row correspond to a non-missing entry in the $m^{th}$ column of $X$.
The marginalized likelihood (\ref{eq:margW}) can be written as
\begin{eqnarray}
\hspace{0.7cm} p\left( X \given U \right)
& =  \prod_{m=1}^M ~  \mathcal{N}\left( \mathbf{x}_{n_m, m} \given 0, ~ \sigma_v^2 \Lambda_{un} \right),
\end{eqnarray}
where $\Lambda_{un}= \beta_v \mathbb{I}+\tilde{U}_n\tilde{U}_n^T$ and $\beta_v = \frac{\sigma^2}{\sigma_v^2}$.
The marginalized posterior is given by
\begin{eqnarray}
\label{eq:margPostVMiss}
p (U \given  X  ) \propto   \exp \left\{ -\frac{1}{2 \sigma_u^2} \Tr (UU^T) \right\} \\
 \times \mid \Lambda_{un} \mid^{-M/2} \nonumber
\exp \left\{ -\frac{1}{2}  \mathbf{x}_{n_m, m}^T ~ \Lambda_{un}^{-1} ~ \mathbf{x}_{n_m, m} \right\}~.
\end{eqnarray}
As shown in (\ref{eq:margPostVMiss}), in presence of missing values, the posterior cannot be factorized as in (\ref{eq:factLam}) because each column $\mathbf{x}_{:m}$ contributes to different blocks $\Lambda_{un}$ of $\Lambda$. 

We propose to address the missing value issue by gap-filling. In particular, if one can obtain a good estimate of the covariance structure in $X$, so that $\Psi_u = \frac{1}{\sigma_v^2} XX^T$ in \myref{eq: margPostV} can be approximated well, one can use the $\mathcal{MGIG}$ posterior to do approximate inference. We consider two simple approaches to approximate the covariance structure of $X$: (i) by zero-padding the missing value matrix $X$ (assuming $E[X]=0$ or centering the data in practice), and estimating the covariance structure based on the zero-padded matrix, and (ii) by using a suitable matrix completion method, such as PMF, to get point estimates of the missing entries in $X$, and estimating the covariance structure based on the completed matrix. We experiment with both approaches in Section~\ref{sec:res}, and the zero-padded version seems to work quite well. % in practice. %In appendix, we provide guarantees for the estimation accuracy of the covariance based on the zero-padded data matrix. 

\subsection{Collapsed Monte Carlo Inference for PMF}
\label{sec:inference}
\begin{center}
\begin{algorithm}[t]
\caption{ CMC Inference for PMF}
\label{alg:CBPMF}
\begin{algorithmic}[1]
  \algnotext{EndFor}
  \State Construct zero-padded matrix $Z$ from $X \in \mathbb{R}^{N\times M}$.
  \State Let $\Psi_u = \frac{ZZ^T}{\sigma_v^2}$, $\Phi_u = \frac{\mathbb{I}}{\sigma_u^2}$, and $\nu_u = \frac{N-M+1}{2}$.
  \State Solve (\ref{eq:modMGIG}) to find mode $\Lambda^*$ of  $\mathcal{MGIG}(\Psi_u , \Phi_u, \nu_u)$.
  \State Let $L^T L = \Lambda^*$ be the Cholesky factorization of $\Lambda^*$.
  \State Let $\tilde{L} = \frac{L}{\sqrt{\rho-M-1}}$.
%   \Statex \hspace{2cm}
   \For{$t = 1 \cdots T$}   %\Comment{Finding $\mathbb{E}[\mathcal{MGIG}(\Psi , \Phi, \nu)]$}
   \vspace{0.05cm}
  \State Let $\Lambda^{(t)} \sim \mathcal{W}_N(\frac{\Lambda^*}{\rho-M-1}, \rho, \tilde{L})$
\Comment{Algorithm 1}
  %where
%  \Statex  \hspace{1cm}$q(\Lambda) = \mathcal{IW}_N(\rho+N+1) \Lambda^*, \rho, L)$ or
%  \Statex  \hspace{1cm} $q(\Lambda) =\mathcal{W}_N(\frac{\Lambda^*}{\rho-M-1}, \rho, L)$.
  \State \label{st:weight} Let $w^t = \frac{\mathcal{MGIG}_N(\Lambda^{(t)} | \Psi_u , \Phi_u, \nu_u)}
{\mathcal{W}_N(\Lambda^{(t)} | \frac{\Lambda^*}{\rho-M-1}, \rho, \tilde{L})}$.
  %{q(\Lambda^t)}$.
  %$w^t = \frac{\mathcal{MGIG}_N(\Lambda^t | \Psi_u , \Phi_u, \nu_u)}{\mathcal{IW}_N(\Lambda^t | (\rho+N+1) \Lambda^*, \rho)}$.
%  \State Let $\bar{\Lambda} = \bar{\Lambda} + w^t \Lambda^t $.
 % 	\EndFor
 % 	\State $\bar{\Lambda} = \frac{\bar{\Lambda}}{\sum_{t=1}^T w^t}$.
%   \Statex \hspace{2cm}
%   \For{$n = 1 \cdots N$} \Comment{Predictions on $\mathbf{x}^*}
   \State \label{st:mu}  Let $\mu^t = {\Lambda}^{(t)}_{*o} {{\Lambda}^{(t)} _{oo}}^{-1} \mathbf{x}^{o}$.
   \State Let $\Sigma^t ={\Lambda}^{(t)} _{**} - {\Lambda}^{(t)} _{*o} {{\Lambda}^{(t)} _{oo}}^{-1} {\Lambda}^{(t)} _{o*}$.
   \State Let $\bar{\mu} = \bar{\mu} + w^t \mu^t$.
   \State \label{st:sig}  Let $\bar{\Sigma} = \bar{\Sigma} + w^t \Sigma^t $.
 %     \EndFor
      \EndFor

  %     \For{$n = 1 \cdots N$} \Comment{Predictions}
      \State $\tilde{\mu}^* = \frac{\bar{\mu}}{\sum_{t=1}^T w^t}$.
      \State $\tilde{\Sigma}^* = \frac{\bar{\Sigma}}{\sum_{t=1}^T w^t}$.
      	\State Report the distribution of $\mathbf{x}^{*} \sim \mathcal{N} (\tilde{\mu}^*, \tilde{\Sigma}^*)$.
	\State Set the point estimate of $\mathbf{x}^{*}$ as $\mu^*$.
%      \EndFor

\end{algorithmic}
\end{algorithm}
\end{center}

Given that ${\Lambda}_u \sim \mathcal{MGIG}_N$, we predict the missing values as follows. Let $\mathbf{x} = [\mathbf{x}^{o}, \mathbf{x}^{*}] \sim \mathcal{N} (0, {\Lambda})$, where $\mathbf{x}^o \in \mathbb{R}^p$ is the observed partition of $\mathbf{x} \in \mathbb{R}^N$ and $\mathbf{x}^* \in \mathbb{R}^{N-p}$ is missing. Accordingly, partition ${\Lambda}$ as
%\begin{align}
%{\Lambda} = \left[ \begin{array}{cc}
%{\Lambda}_{oo} & {\Lambda}_{o*} \\
%{\Lambda}_{*o} & {\Lambda}_{**}
%\end{array} \right] \notag
%\end{align}
%\vskip -12em
\begin{align}
\setlength{\abovedisplayskip}{0pt}
\setlength{\belowdisplayskip}{0pt}
{\Lambda}_u = \begin{blockarray}{ccc}
p & N-p \\
\begin{block}{(cc)c}
  {\Lambda}_{oo} & {\Lambda}_{o*}  & p\\
{\Lambda}_{*o} & {\Lambda}_{**} & N-p \\
\end{block}
\end{blockarray}.
\end{align}
%\vspace{-0.5cm}
\noindent Then the conditional probability of $\mathbf{x}^{*}$ given $\mathbf{x}^{o}$ and $\Lambda$ is
\begin{align}
\label{eq:condNorm}
p(\mathbf{x}^{*} \given \mathbf{x}^{o}, \Lambda) & \sim \mathcal{N} (\mu^*, \Sigma^*),  \\
\mu^*  &= \Lambda_{*o} \Lambda_{oo}^{-1} \mathbf{x}^{o}, \notag \\
\Sigma^*  &= \Lambda_{**} - \Lambda_{*o} \Lambda_{oo}^{-1} \Lambda_{*o}. \notag
\end{align}
where $\mathbf{y} = \Lambda_{*o} \Lambda_{oo}^{-1}$ is the solution of the linear system $\Lambda_{oo} \mathbf{y} = \Lambda_{*o}^T$ and can be calculated efficiently.
%where $\mu^*$ gives the point estimate of the missing entries and (\ref{eq:condNorm}) provide the distribution over the missing entries.
Since sampling from $\mathcal{MGIG}$ is difficult, we propose to use importance sampling to infer the missing values as
\begin{align*}
p(x_n^* | x_n^o )= \mathbb{E}_{\Lambda \sim \mathcal{MGIG}} \left[ p(x_n^* | x_n^o , \Lambda) \right]  = \mathbb{E}_{\Lambda\sim q} \left[ \frac{p(x_n^* | x_n^o , \Lambda) \mathcal{MGIG}_N(\Lambda | \Psi_u, \Phi_u, \nu_u)}{q(\Lambda)} \right],
\end{align*}
where $q$ is the proposal distribution as discussed above and sampling $\Lambda^{(t)}$ from $q$ yields to the estimate of
\begin{align}
\tilde{\mu}^*  &=  \frac{\sum_{t=1}^T \Lambda^{(t)}_{*o} {\Lambda^{(t)}_{oo}}^{-1} \mathbf{x}^{o} w(\Lambda^{(t)})}{\sum_{t=1}^t w(\Lambda^{(t)})} \notag \\
%\mathbb{E}_{\Lambda\sim q} \left[ \right] \notag \\
\tilde{\Sigma}^*  &= \frac{\sum_{t=1}^T [\Lambda^{(t)}_{**} - \Lambda^{(t)}_{*o} {\Lambda^{(t)}_{oo}}^{-1} \Lambda^{(t)}_{*o}] w(\Lambda^{(t)})}{\sum_{t=1}^t w(\Lambda^{(t)})}.
\end{align}
%\begin{figure*}
%\begin{center}
%\subfigure[LL percentile MovieLens]{\includegraphics[width = 0.32\textwidth]{Figures/movieRatingRemovedRatLess10_Perplexity_Percentile.eps}
%\label{fig:perctileMovie}}
%\subfigure[LL percentile SNP]{\includegraphics[width = 0.32\textwidth]{Figures/genotypes_chr13_CEU_r21_nr_fwd_phased_k10.eps}
%\label{fig:perctileSNP}}
%\subfigure[LL percentile Gene Expression]{\includegraphics[width = 0.32\textwidth]{Figures/GeneExpLevel3K10.eps}\label{fig:perctileGene}}
%\end{center}
%\vspace{-0.5cm}
%\caption{Log loss (LL) of CMC and MCMC for different LL percentile on  s) MovieLens data, b) SNP, and c) Gene Expression presented in the log scale. log loss of MCMC is increasing exponentially (linear in log scale) by adding data points with higher log loss.}
%\label{fig:perctileAll}
%\vspace{-0.5cm}
%\end{figure*}

Algorithm \ref{alg:CBPMF} in appendix illustrates the summary of the collapsed Monte Carlo (CMC) inference for predicting the missing values.
A practical approximation to avoid the calculations in Lines 9-12 of Algorithm \ref{alg:CBPMF} at each iteration is to simply estimate the mean of the posterior $\bar{\Lambda} = \frac{\sum_{t=1}^T \Lambda^{(t)} w^t}{\sum_{t=1}^T w^t} $ with samples drawn from the proposal distribution (line 6), then do the inference based on $\bar{\Lambda}$.
As it is shown in Section \ref{sec:res}, if the degrees of freedom $\nu_u$ is small,  the mode is close to the mean and the approximation using $\bar{\Lambda}$ works well.

\section{Experimental Results}
\label{sec:res}
We compared the performance of MCMC and CMC on both log loss and running times.

\subsection{Datasets}
We evaluated the models on 4 datasets:
%\noindent \textbf{SNP:}
1) \textbf{SNP}: single nucleotide polymorphism (SNP) is important for identifying gene-disease associations where the data usually has 5 to 20\% of genotypes missing~\cite{dahm04}.
%A minor allele frequency, the lowest allele frequency at a locus that is observed in a particular population, is denoted by 1, and the majority alleles are denoted by 0.
We used phased SNP dataset for chromosome 13 of the CEU population\footnote{http://hapmap.ncbi.nlm.nih.gov/downloads/phasing/}.
We randomly dropped 20\% of the entries.
%\noindent \textbf{Gene Expression:}
2) \textbf{Gene Expression:} DNA microarrays provides measurement of thousands of genes under a certain experimental condition where suspicious values are usually regarded as missing values.
%high throughput investigation of gene expressions by simultaneously measuring the expression of thousands of genes under a certain experimental condition.
%Suspicious values are usually regarded as missing values to avoid affecting the further analyses.
Here we used gene expression dataset for Breast Cancer (BRCA)\footnote{http://cancergenome.nih.gov/}. 
We randomly dropped 20\% of the entries.
% Here we used gene expression dataset for Breast Cancer (BRCA) generated by the TCGA Research Network\footnote{http://cancergenome.nih.gov/} with $M=17,814$ genes and $N=591$ samples. 
% We randomly hide 20\% of entries.
%\noindent \textbf{MovieLens:}
3) \textbf{MovieLens:} we used MovieLens\footnote{www.movielens.umn.edu} dataset with 1M rating represented as a fat matrix $X \in \mathbb{R}^{N \times M}$ where $M=3900$ movies and $N=6040$ users. %Movies with less than 10 ratings are removed yielding to $M=3233$ movies.
%We randomly selected 100 movies among 3900 movies in the dataset. The used subset of MovieLens contains 83,000 ratings meaning 86\% of the ratings are missing.
%\noindent \textbf
4) \textbf{Synthetic:} first the latent matrices $U$ and $V$ are generated by randomly choosing each $\{\mathbf{u}_n\}_{n=1}^N$ and $\{\mathbf{v}_m\}_{m=1}^M$ from $\mathcal{N}(0, \sigma_u^2 \mathbb{I})$ and $\mathcal{N} (0, \sigma_v^2 \mathbb{I})$, respectively. Then, matrix $X$ is built by sampling each $x_{nm}$ from $\mathcal{N} ( \langle \mathbf{u}_n, \mathbf{v}_m \rangle, \sigma^2)$.
The parameters are set to $N=100$, $M=6000$, $\sigma_u^2=\sigma_v^2=0.05$, and $\sigma^2 = 0.01$. We dropped random entries using Bernoulli distributions with $\delta = 0.1, 0.2$.
\subsection{Methodology}
We compared CMC with MCMC inference for PMF. Gibbs sampling with diagonal covariance prior over the latent matrices is used for MCMC.
%\textbf{RMSE:} The most common evaluation measure for prediction accuracy is the root of the mean square error (RMSE), given as $RMSE = \frac{1}{T} \sqrt{\sum_{i=1}^N \sum_{j=1}^M \delta_{ij} (x_{ij} - \hat{x}_{ij})^2}$
%where $x_{ij}$ is the true rating , $\hat{x}_{ij}$ is the predicted value for user $i$ and movie $j$, and $T$ is the total number of ratings .
%
For the model evaluation, average of log loss (LL) is reported over 5-fold cross-validation.
LL measures how well a  probabilistic model $q$ predicts the test sample defined as $LL = {-\frac{1}{T} \sum_{i=1}^N \sum_{j=1}^M \delta_{ij} \log q(x_{ij})}$ where $q(x_{ij})$ is the inferred probability and $T$ is the total number of observed values. A better model $q$ assign higher probability $q(x_{ij})$ to observed test data, and have a smaller value of LL.
% For evaluating probabilistic models which infer a distribution over missing values, it makes more sense to use measures such as log-loss or perplexity which considers the shape of the distribution and uncertainty in predictions rather than RMSE which only captures proximity to the mean~\cite{hehg14}.

\noindent {\bf LL Percentile:} For any posterior model $q(x)$, a test data point $x_{\text{test}}$ with low $q(x_{\text{test}})$ has large log loss, and high $q(x_{\text{test}})$ has low log loss. To comparatively evaluate the posteriors obtained from MCMC and CMC, we consider their log loss percentile plots.
For any posterior, we sort all the test data points in ascending order of their log loss, and plot the mean log loss in 10 percentile batches. More specifically, the first batch corresponds to the top 10\% of data points with the lowest log loss, the second batch corresponds to the top 20\% of data points with the lowest log loss (including the first 10\% percentile), and so on. %For the percentile plots, we consider the mean log loss in each batch.

%{\bf AB: This part can be dropped - folks understand this.} To illustrate it better, consider a probabilistic model where the mean of predicting distribution is close to the true value (low RMSE) but with a low probability of generating the true value (large log loss) due to a low standard deviation (Figure \ref{fig:densityMovie3} - blue curve). This model is risky because the model infers the ratings with high confidence (low standard deviation), and low RMSE  but with zero chance of happening of the true value.
\subsection{Results}
\begin{figure}[t]
\centering
\subfigure[Synthetic with $\delta = 0$]{\includegraphics[width = 0.3\textwidth]{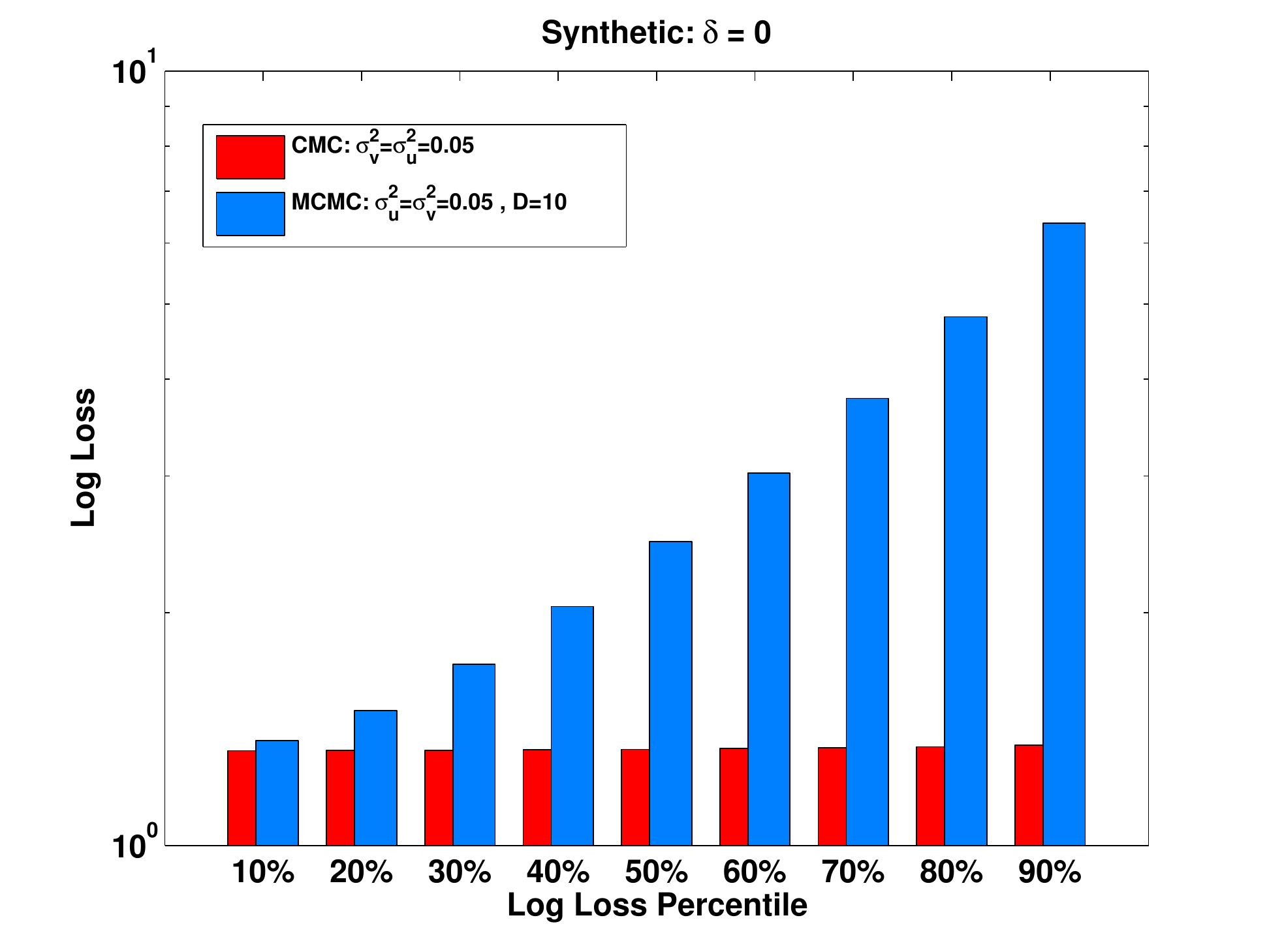}
\label{fig:perctileSynd0}}
\subfigure[Synthetic with $\delta = 0.1$]{\includegraphics[width = 0.3\textwidth]{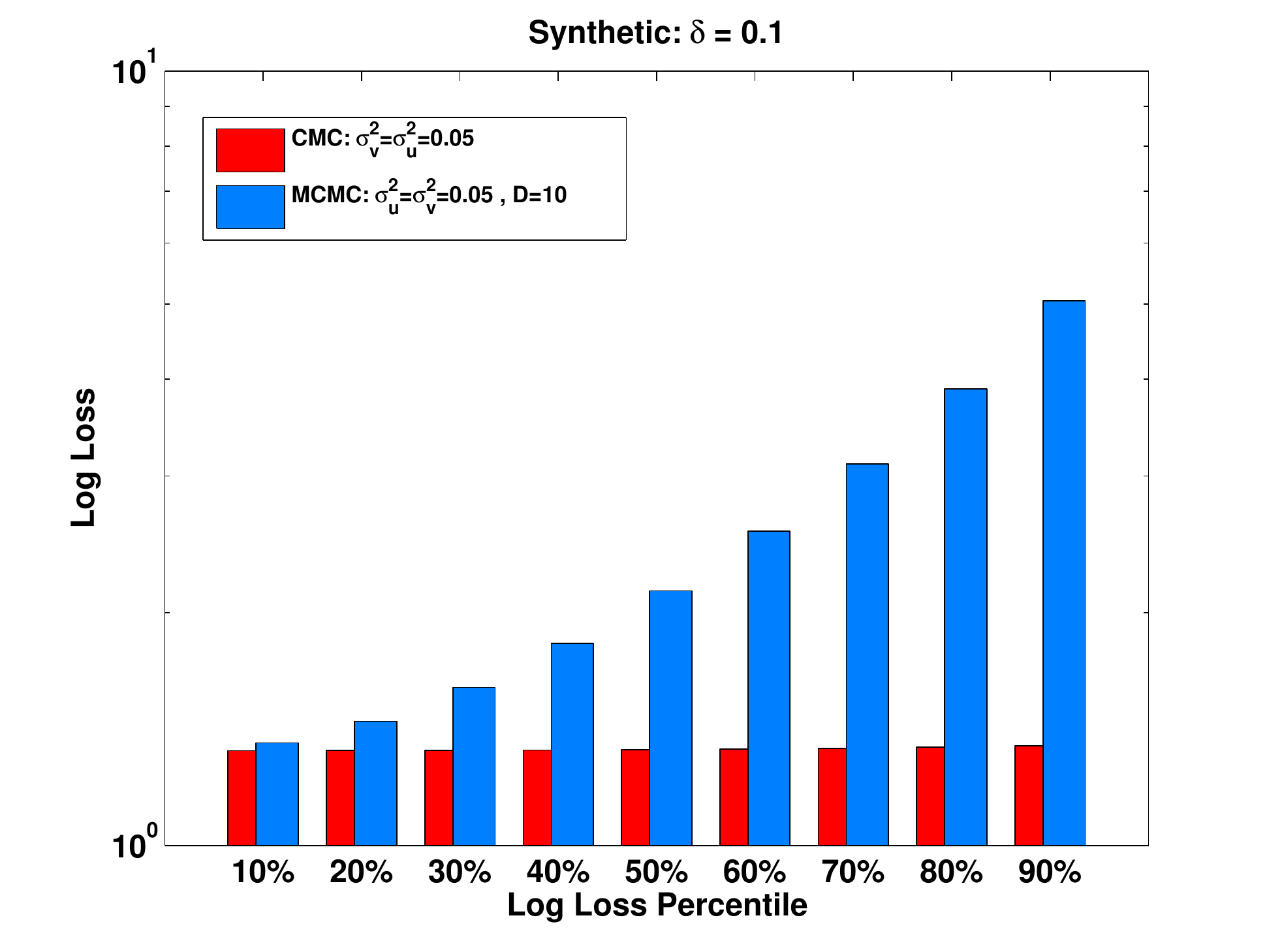}
\label{fig:perctileSyndp1}}
\subfigure[Synthetic with $\delta = 0.2$]{\includegraphics[width = 0.3\textwidth]{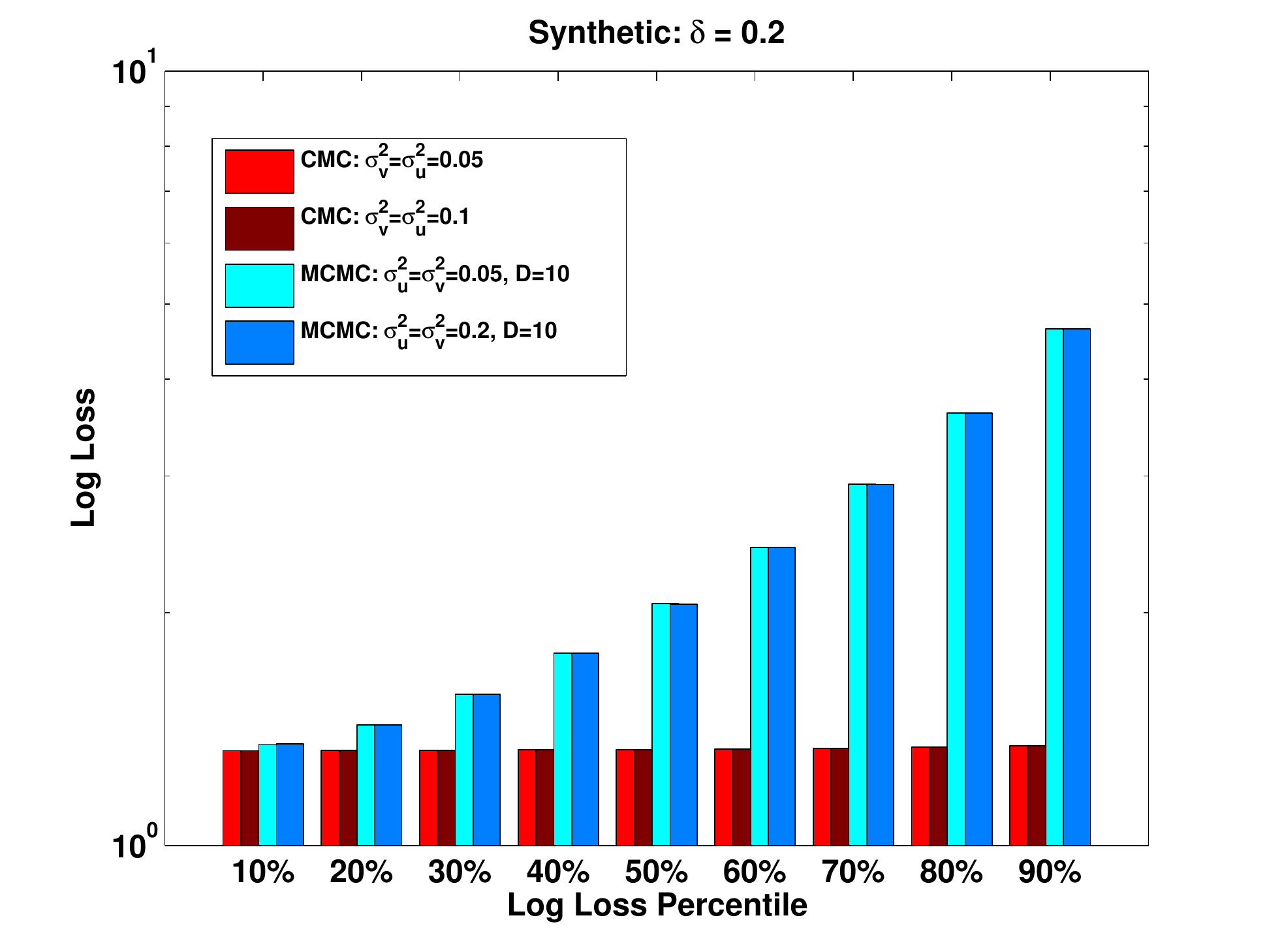}
\label{fig:perctileSyndp2}}
\subfigure[MovieLens]{\includegraphics[width = 0.3\textwidth]
{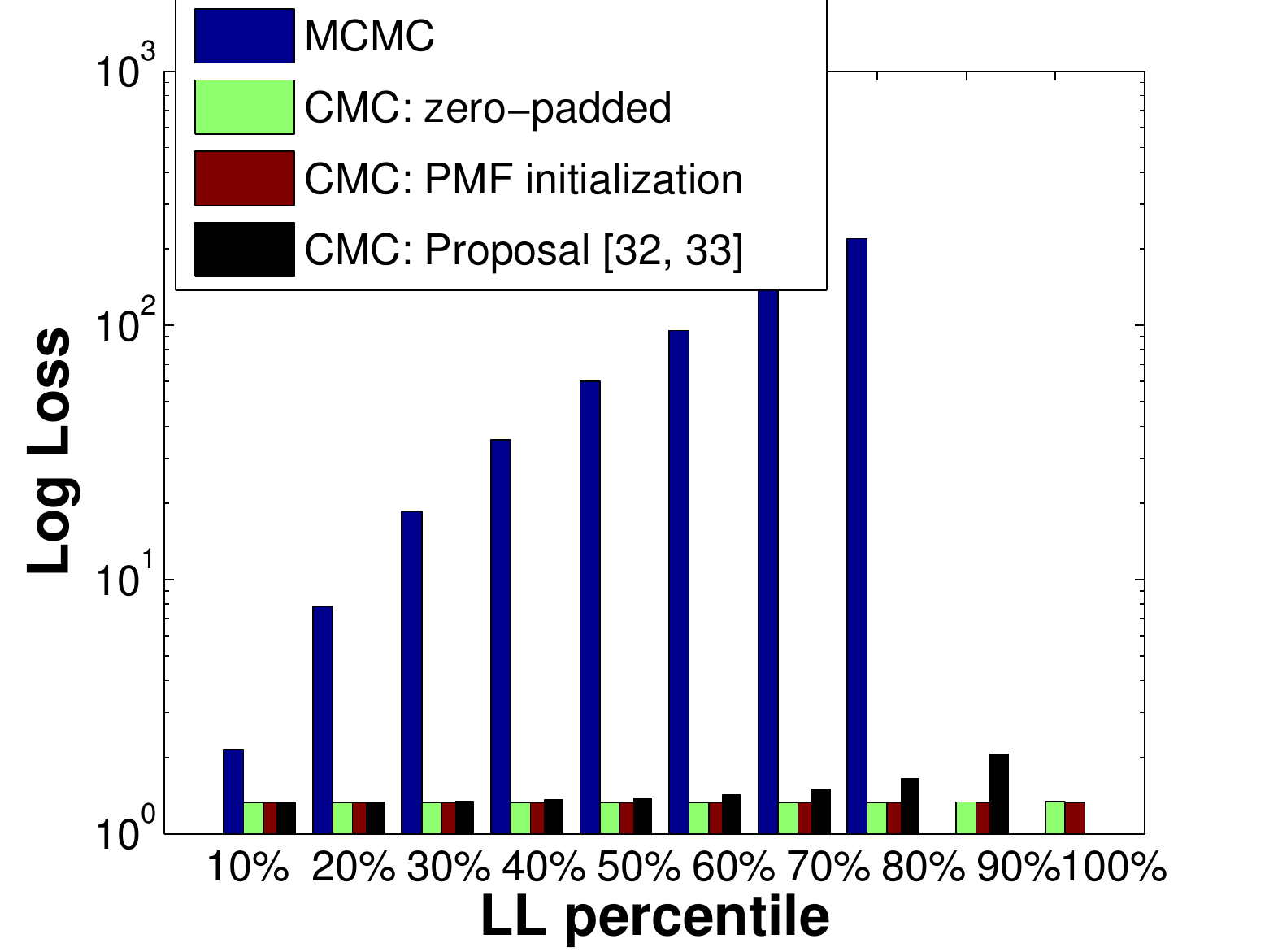}
%{movieRatingRemovedRatLess10_Perplexity_Percentile-eps-converted-to.pdf}
\label{fig:perctileMovie}}
\subfigure[SNP]{\includegraphics[width = 0.3\textwidth]
%{genotypes_chr13_CEU_r21_nr_fwd_phased_k10-eps-converted-to.pdf}
{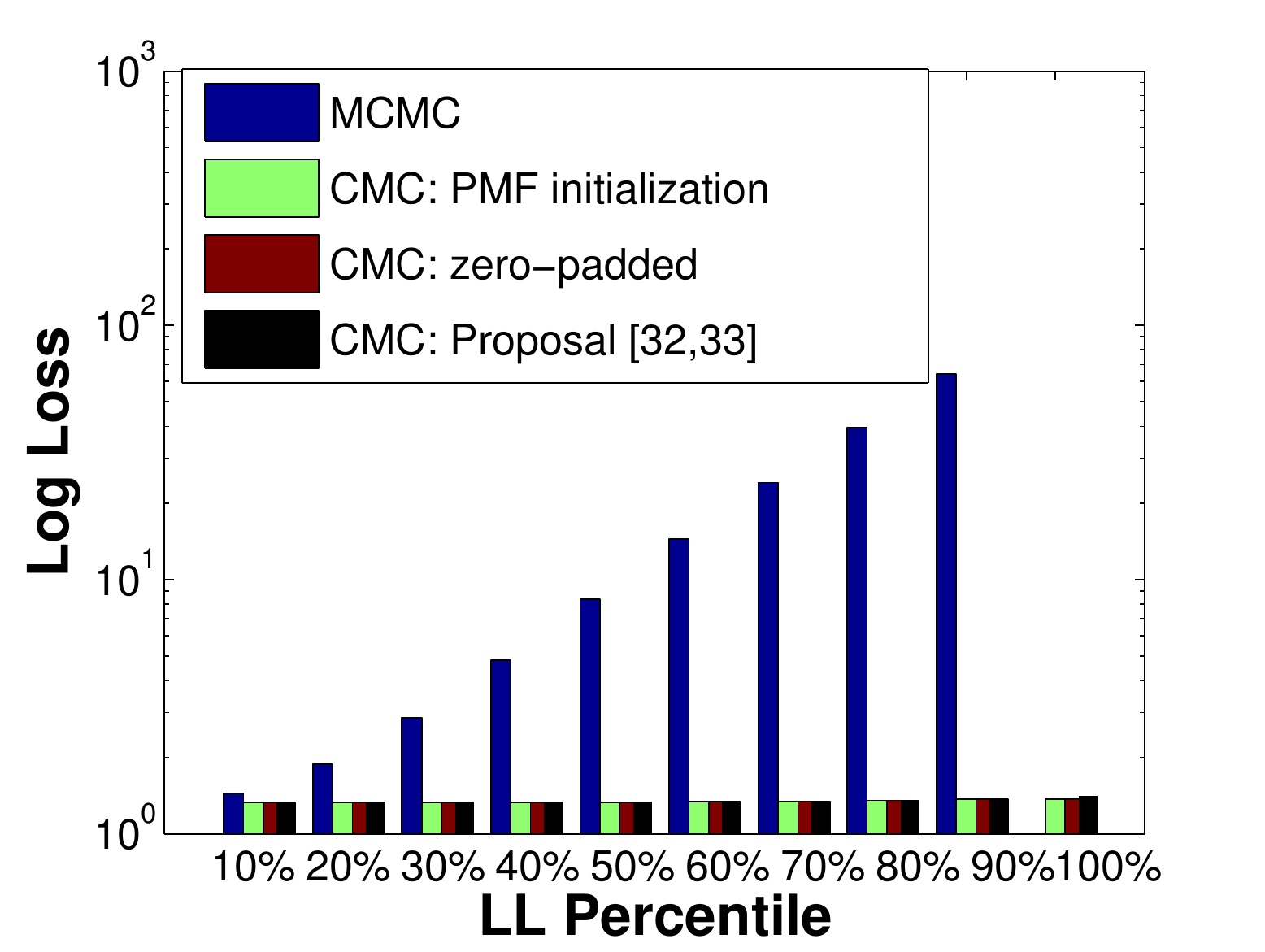}
\label{fig:perctileSNP}}
\subfigure[Gene Expression]{\includegraphics[width = 0.3\textwidth]
%{GeneExpLevel3K10-eps-converted-to.pdf}
{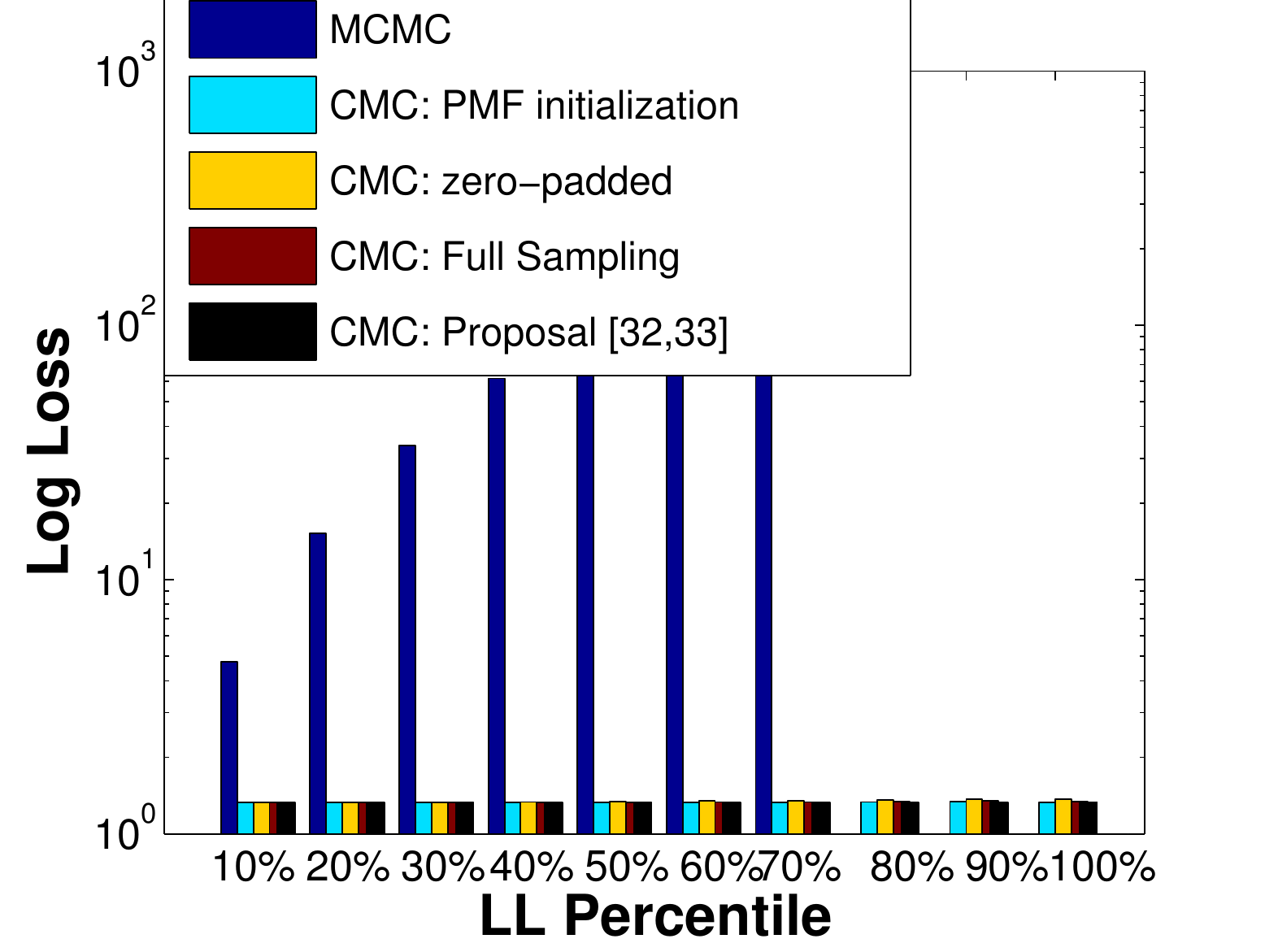}
\label{fig:perctileGene}}
%\subfigure[$\delta = 0$]{\includegraphics[width = 0.3\textwidth]{rmse_synthetic_n6000_d100_sigUp05_sigVp05_nSam100_delta0-eps-converted-to.pdf}
%\label{fig:rmseSynd0}}
%\subfigure[$\delta = 0.1$]{\includegraphics[width = 0.3\textwidth]{rmse_synthetic_n6000_d100_sigUp05_sigVp05_nSam100_deltap1-eps-converted-to.pdf}
%\label{fig:rmseSyndp1}}
%\subfigure[$\delta = 0.2$]{\includegraphics[width = 0.3\textwidth]{rmse_synthetic_n6000_d100_sigUp05_sigVp05_nSam100_deltap2_diffparams-eps-converted-to.pdf}
%\label{fig:rmseSyndp2}}
\caption{Log loss (LL) of CMC and MCMC for different log loss percentile on different datasets presented in the log scale ($\delta$ denotes the missing proportion). CMC consistently achieves lower LL compared to MCMC. LL of MCMC increases exponentially (linearly in log scale) by adding data points with higher log loss. Proposal in [30,31] achieved infinity LL for MovieLens. Empty bar represents infinity LL (e.g. 90\% and 100\% percentile in (d)}
%(c-e) RMSE of CMC and MCMC for different number of samples on synthetic data . MCMC achieves lower RMSE compared to CMC with fewer samples but having higher log loss than CMC.}
\label{fig:perctileSyn}
\end{figure}

\begin{figure}[t]
\begin{center}
\subfigure[LL of MCMC]{\includegraphics[width = 0.45\textwidth]{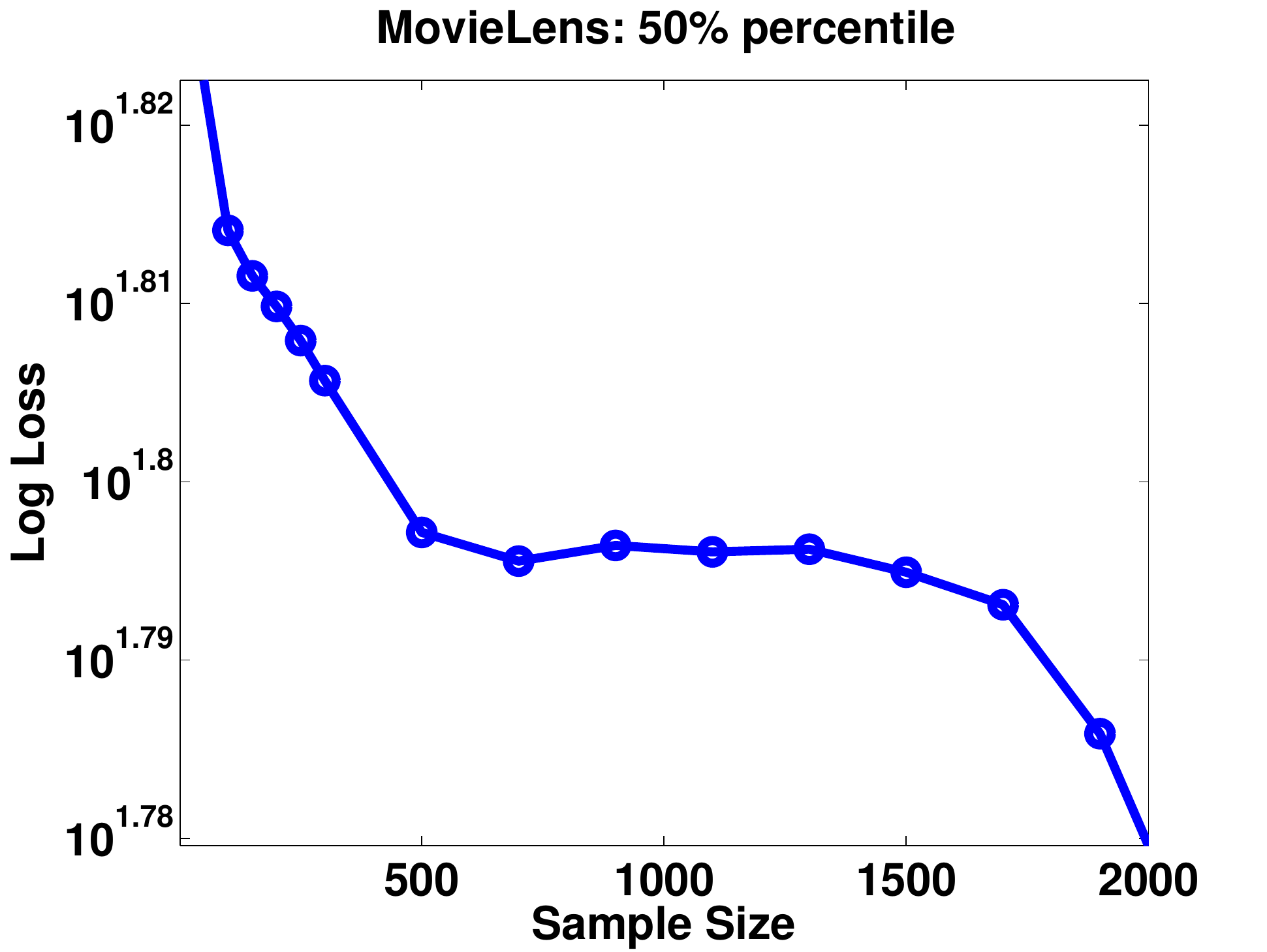}\label{fig:lossMovieInd1}}
\subfigure[LL of CMC]{\includegraphics[width = 0.45\textwidth]{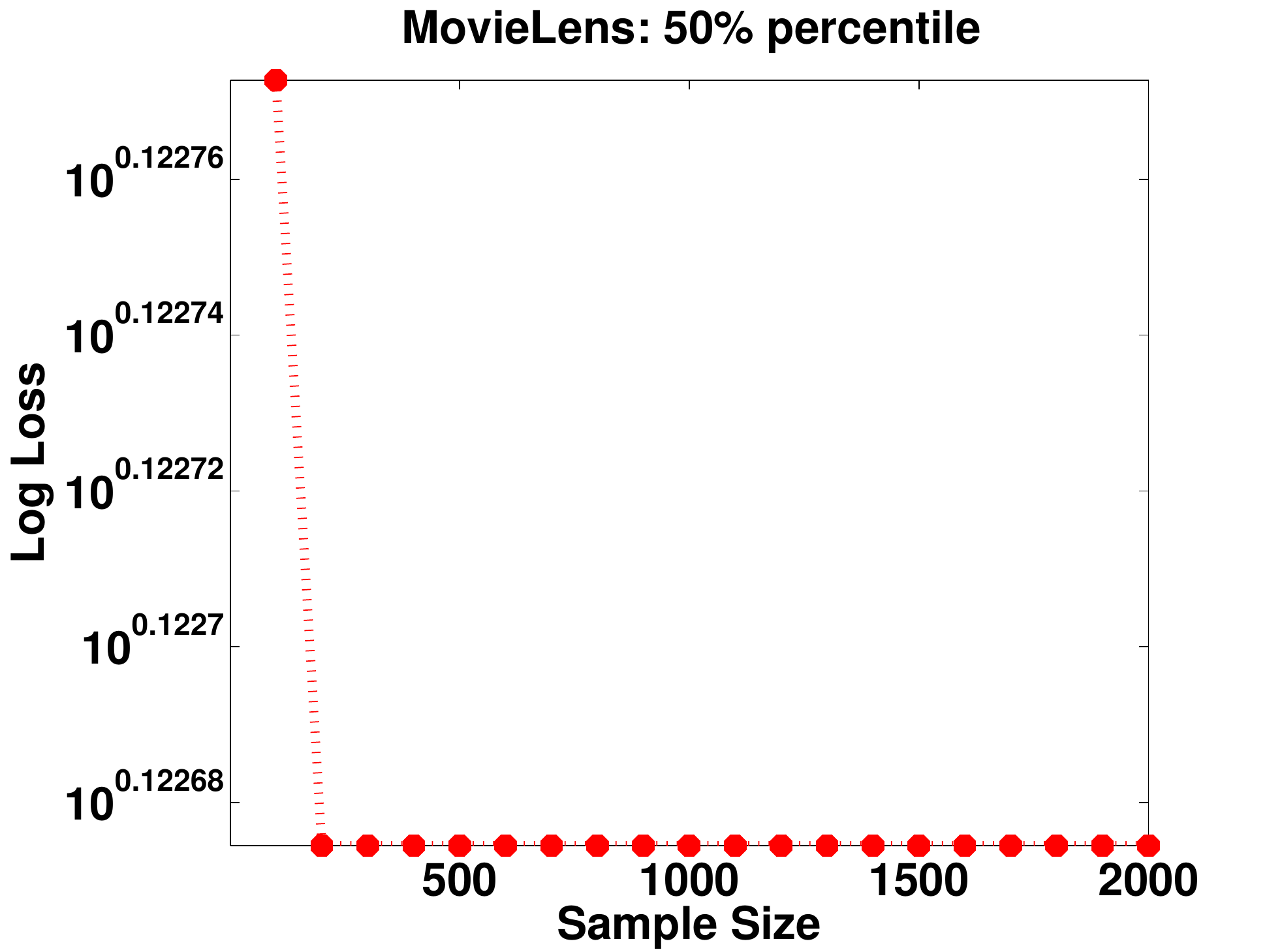}\label{fig:lossMovieInd2}}
\end{center}
\caption{
LL of CMC and MCMC for different sample size of MovieLens data in the log scale. LL of both CMC and MCMC is decreasing by adding more samples. LL of MCMC is in magnitude 10 times more than CMC.
}
\label{fig:lossIteration}
\end{figure}

We summarize the results from different aspects:
%First, the effective number of samples used in CMC and MCMC is discussed. Next, we comparatively evaluate  CMC and MCMC based on log loss and RMSE. Finally, the inferred posterior distributions from CMC and MCMC are compared.

\noindent \textbf{Log loss:}
\label{sec:resLogloss}
CMC has a small log loss across all percentile batches, whereas log loss of MCMC increases exponentially (linear increase in the log scale) for percentile batches with  higher log loss i.e., smaller predicting probability, (Figure \ref{fig:perctileSyn}). Thus, MCMC assigned extremely low probability to several test points as compared to CMC.
%Figures shows the log loss percentile plot on MovieLens, SNP, and gene expression data. Similar to the synthetic data, CMC provides smaller log loss whereas log loss of MCMC increases exponentially (linear in the log scale) by adding data points with higher log loss. %It also can be observed that MCMC is more sensitive to changing the parameters comparing to CMC.
Figure \ref{fig:lossMovieInd1} illustrates that log loss of MCMC continues to decrease with growing sample size up to 2000 samples, implying that MCMC has not yet converged to the equilibrium distribution. Note that log loss of CMC with 200 samples (Figure \ref{fig:lossMovieInd2}) is 10 times less than log loss of MCMC with 2000 samples.
We also compared the results with the previous proposal~\cite{yalz13,yoshii13}, and observed that for MovieLens the results are worse than our proposed result as they achieved Inf LL on the last batch.

\noindent \textbf{Effective number of samples:}
\label{sec:sample}
For the synthetic, SNP, and gene expression datasets, we generated 10,000 samples using MCMC.
The burn-in period is set to 500 with a lag of 10 yielding to 1000 effective samples.
For the MovieLens, we generated 5,000 samples using MCMC with
the burn-in period of 1000 and a lag of 2 yielding to 2000 effective samples.
We initialized the latent matrices $U$ and $V$  with the factors estimated by PMF, to help the convergence of MCMC.
Sample size in CMC procedure is set to 1,000 for all datasets.
Note that MCMC alternately sample both latent matrices $U$ and $V$ from a Markov chain and the quality of the posterior improves with increasing number of samples.
For the proposed CMC procedure, the bigger matrix $V$ is marginalized and only samples from the smaller $U$ matrix is drawn directly from the true posterior distribution.
Hence, CMC has considerably improved sample utilization.
%In CMC, a Monte Carlo sampling from the posterior of only one of the latent matrix is applied, unlike MCMC which alternately sample both latent matrices.
%The proposed CMC procedure draw samples directly from the posterior, whereas MCMC draw samples from a Markov chain that has the desired distribution as its equilibrium distribution.
%Hence MCMC significantly improves its sample utilization.
%

\noindent \textbf{Initialization:}
As discussed in Section \ref{sec:missPostPMF}, in order to use the $\mathcal{MGIG}$ posterior for  inference, the covariance structure of matrix $X$ should be estimated.
Here we evaluate two approaches to approximate the covariance structure of X: (i) by zero-padding the missing value matrix X, and (ii) by computing the point estimates of the missing entries in X with PMF. CMC with zero-padded initialization has a similar log loss behavior as point estimate initialization with PMF (Figures \ref{fig:perctileSyn} (d-f)).

\noindent \textbf{Full sampler vs Mean sampler:}
Figure \ref{fig:perctileGene} shows the result of the full sampler (Algorithm \ref{alg:CBPMF}), and the mean sampler (approximating the inference by estimating $\bar{\Lambda} = \mathbb{E}_{\Lambda \sim \mathcal{MGIG}} [\Lambda]$ as discussed in Section \ref{sec:inference}) on gene expression data. Since the log losses are similar with both samplers, and the behavior is typical, we presented log loss results on the other datasets only based on the mean sampler, which is around 100 times faster.

\begin{figure}[t]
\centering
\subfigure[{\scriptsize{LL- CMC:-1.78, MCMC:-Inf}}]{\includegraphics[width = 0.24\textwidth]{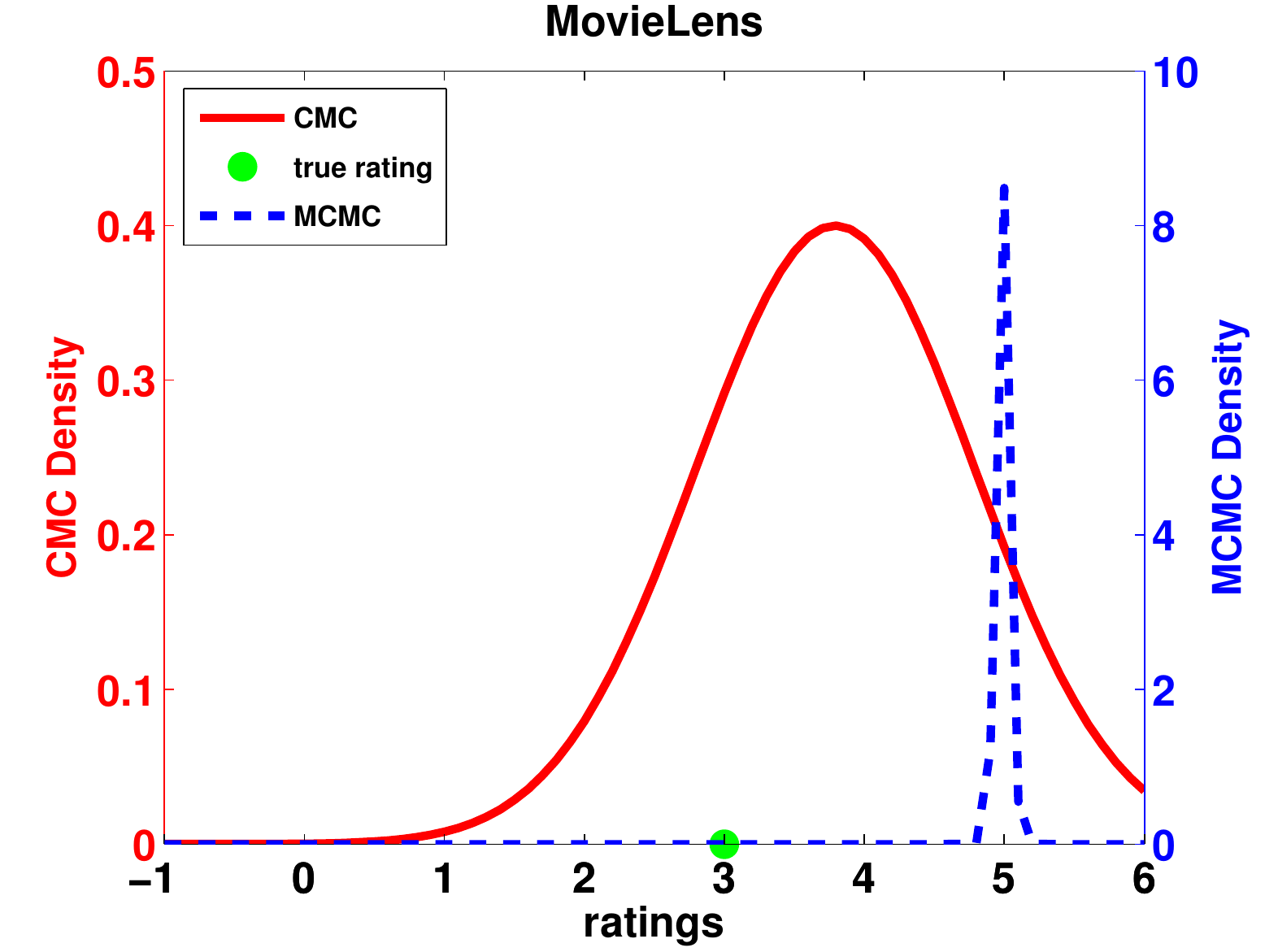}
\label{fig:densityMovie1}}
\hspace{-0.3cm}
\subfigure[{\scriptsize{LL- CMC:-3, MCMC:-17}}]{\includegraphics[width = 0.24\textwidth]{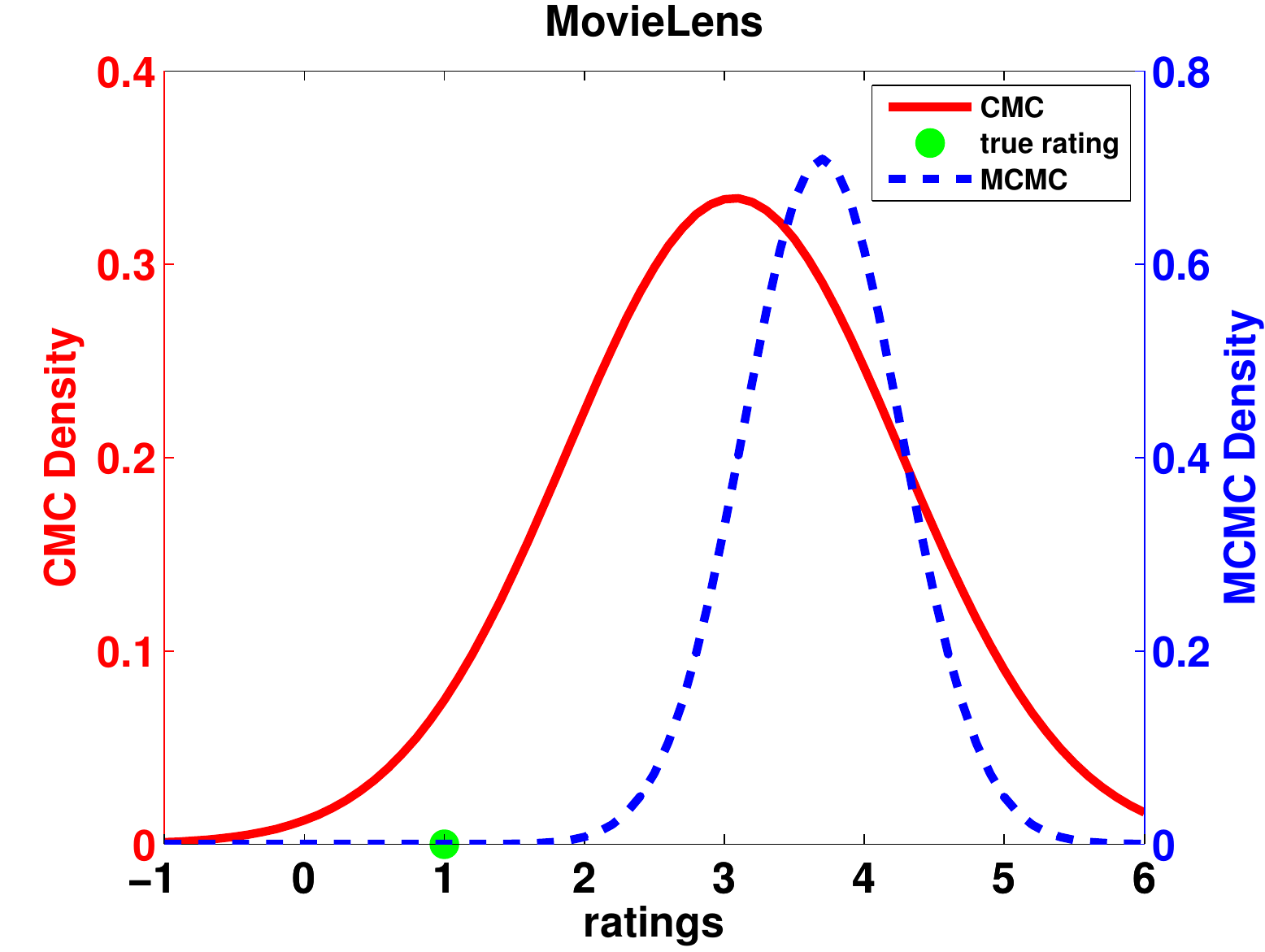}
\label{fig:densityMovie2}}
\hspace{-0.3cm}
\subfigure[{\scriptsize{LL- CMC:-4.2, MCMC:-6.4}}]{\includegraphics[width = 0.24\textwidth]{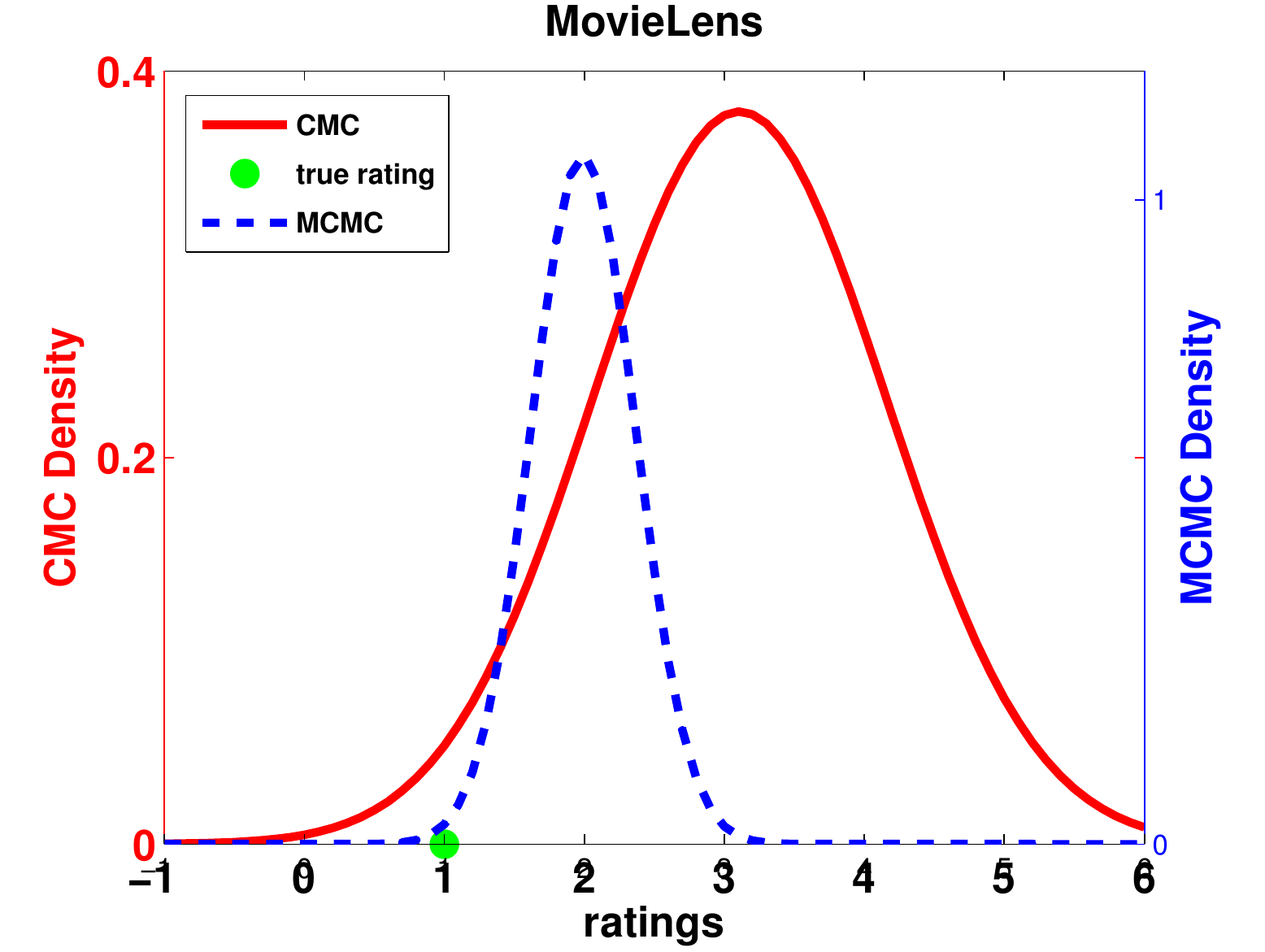}
\label{fig:densityMovie3}}
\hspace{-0.3cm}
\subfigure[{\scriptsize{LL- CMC:-1.4, MCMC:--2.04}}]{\includegraphics[width = 0.24\textwidth]{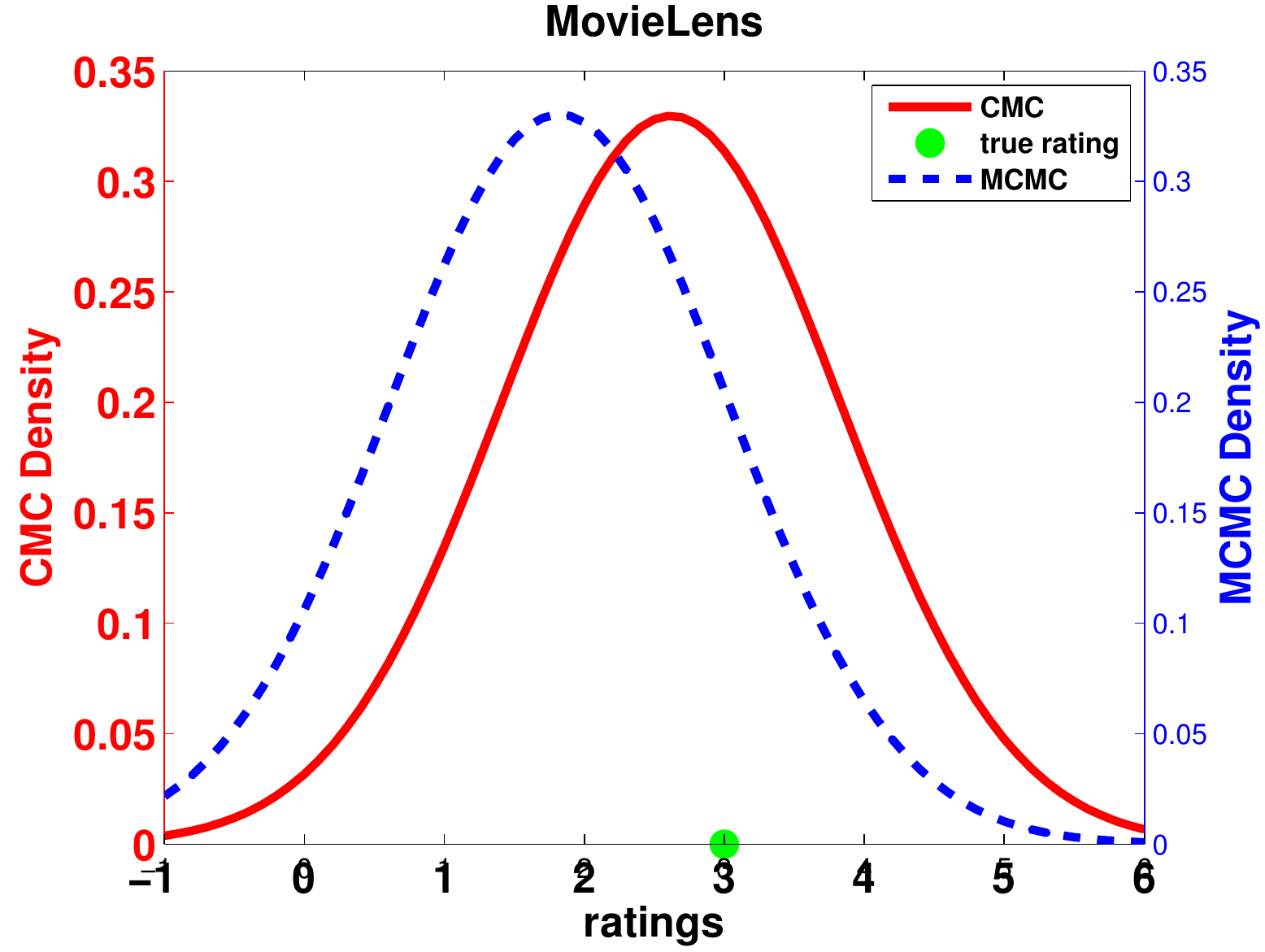}
\label{fig:densityMovie3}}
\caption{Density of CMC and MCMC for several data input on  MovieLens data. CMC provide distributions with lower LL compared to MCMC e.g. in (a) LL of MCMC is -Inf whereas LL of CMC is -1.78.}
\label{fig:densityMovie}
\end{figure}

\noindent \textbf{Comparison of inferred posterior distributions:}
\label{sec:resDens}
To emphasize the importance of choosing the right measure for comparison, e.g., log loss vs RMSE, we illustrate the inferred posterior distributions over several missing entries/ratings in MovieLens obtained from MCMC and CMC in Figure \ref{fig:densityMovie}. Note that the scales for CMC (red) and MCMC (blue) are different. Overall, the posterior from CMC tends to be more conservative (not highly peaked), and obtains lower log loss across a range of test points. Interestingly, as shown in  Figure \ref{fig:densityMovie1}, MCMC can make mistakes with high confidence, i.e., predicts 5 stars with a peaked posterior whereas the true rating is 3 stars. Such troublesome behavior is correctly assessed with log loss, but not by RMSE since it does not consider the confidence in the prediction.
%log loss of MCMC is -Inf whereas LL of CMC is -1.78. Here, MCMC provide a distribution with high confidence (low standard deviation) but with zero true probability.
As shown in Figure \ref{fig:densityMovie3}, for some test points, both MCMC and CMC inferred similar posterior distributions with a bias difference where the mean of CMC is closer to the true value.

\noindent \textbf{Time Comparison:}
We have compared running time in both serial and in parallel over 1000 steps yielding to 200 and 1000 samples for MCMC and CMC, respectively. We implement the algorithms in Matlab.
The computation time is estimated on a PC with a 3.40 GHz Quad core CPU and 16.0G memory.
The average run time results are reported in Table \ref{tab:time}.
For Synthetic, SNP, and gene expression datasets, MCMC converges very slowly. For MovieLens dataset, the running time of both are very close but note that MCMC requires more number of samples for convergence than CMC (Figure \ref{fig:lossIteration}).

\begin{table}[t]
\centering
\caption{Time Comparison of CMC and MCMC on different datasets. 
At each step of MCMC, rows of $U$ and $V$ can be sampled in parallel denoted by MCMC parallel.
The running time is reported over 1000 steps for both methods. Note that the effective number of samples of MCMC is less than 1000 and more steps is required to obtain enough samples. The number of iterations for convergence of CMC is much less than 1000 (Figure \ref{fig:lossIteration}).}
\label{tab:time}
\begin{tabular}{l  r@{}  c @{} r     r rrr   rr r r   }
\textbf{Dataset} & \multicolumn{3}{c}{\bf{Size}} & \multicolumn{4}{c}{\bf{MCMC (1000 steps, 200 samples)}} & \multicolumn{4}{c}{\bf{CMC (1000 steps, 1000 samples)}} \\
& & & && \bf{Serial} & \bf{Parallel} && &\bf{Serial} &  \bf{Parallel}& \\
\hline
\bf{Synthetic} & 100 & $\times$ & 6,000            && 728s & 404s &&   &6s & 4s &\\
\bf{SNP} &  120 & $\times$ &  104,868            && 12,862s & 5,859s  &&& 75s & 22s &\\
\bf{Gene Expression} & 591  & $\times$ & 17,814 && 3,478s & 2,278s &&& 140s & 90s &\\
\bf{MovieLens} &  3,233 & $\times$&  6,040       && 2,350s & 2,100s && &5,387s & 2,058s &\\
%GeneExp & 20,000s (333m) & 12000s (200m) & 4727s (78.8m) & 456s (7.6m) \\
%SNP & 2492s & 1185s & 40.67s & 8.7s \\
\hline
\end{tabular}
\end{table}

%{\color{red}
%\textbf{5) Complexity:}
%\begin{itemize}
%\item MCMC: $O(S (N+M)D^3)$ \\
%where $S$ is number of generated samples, not effective sample size
%\item CMC: $O(S M^2)$ \\
%where $S$ is the number of $\mathcal{MGIG}$ samples generated from the proposal.
%\end{itemize}
%As an example, I generated a synthetic data with $\# rows:100,000$ , $\# cols:100$. \\
%CMC: Each iteration (Sampling Lambda and Prediction) is taking 53sec and therefore for 50 iterations is taking around 44min. \\
%MCMC: Each iteration (Sampling U,V and Prediction) is taking 4sec and therefore for 10,000 iterations is taking around 11 hours!
%}
%\\\\

\section{Conclusion}
\label{sec:concl}
We studied the $\mathcal{MGIG}$ distribution and provided certain key properties with a novel sampling technique from the distribution and its connection with the latent factor models such as PMF or BPCA.
With showing that the $\mathcal{MGIG}$ distribution is unimodal and the  mode can be obtained by solving an ARE, we proposed an new importance sampling approach to infer the mean of $\mathcal{MGIG}$. The new sampler, unlike the existing sampler~\cite{yalz13,yoshii13}, chooses the proposal distribution to have the same mode as the $\mathcal{MGIG}$. This characterization leads to a far more effective sampler than~\cite{yalz13,yoshii13} since the new sampler align the shape of the proposal to the target distribution. Although, the $\mathcal{MGIG}$ distribution has been recently applied to Bayesian models as the prior for the covariance matrix, here, we introduced a novel application of the $\mathcal{MGIG}$ in PMF or BPCA. We showed that the posterior distribution in PMF or BPCA with Gaussian priors, has the $\mathcal{MGIG}$ distribution. This illustration, yields to a new CMC inference algorithm for PMF. 

\section*{Acknowledgment}
The research was supported by NSF grants IIS-1447566, IIS-1447574, IIS-1422557, CCF-1451986, CNS- 1314560, IIS-0953274, IIS-1029711, NASA grant NNX12AQ39A, and gifts from Adobe, IBM, and Yahoo. F. F. acknowledges the support of DDF (2015-2016) from the University of Minnesota.

%\newpage
\bibliographystyle{plain}
\bibliography{references}

\begin{thebibliography}{10}

\bibitem{anmo07}
Brian~DO Anderson and John~B Moore.
\newblock {\em Optimal control: linear quadratic methods}.
\newblock Courier Corporation, 2007.

\bibitem{bai98}
Zhaojun Bai and James Demmel.
\newblock Using the matrix sign function to compute invariant subspaces.
\newblock {\em SIAM Journal on Matrix Analysis and Applications},
  19(1):205--225, 1998.

\bibitem{babj82}
O.~Barndorff-Nielsen, P.~Bl{\ae}sild, et~al.
\newblock Exponential transformation models.
\newblock {\em Proceedings of the Royal Society of London. A. Mathematical and
  Physical Sciences}, 379(1776):41--65, 1982.

\bibitem{bish99a}
C.~M. Bishop.
\newblock Bayesian {PCA}.
\newblock {\em NIPS}, pages 382--388, 1999.

\bibitem{bish99b}
C.~M. Bishop.
\newblock Variational principal components.
\newblock In {\em ICANN}, 1999.

\bibitem{blei10}
D.~Blei, P.~Cook, and M.~Hoffman.
\newblock Bayesian nonparametric matrix factorization for recorded music.
\newblock In {\em ICML}, pages 439--446, 2010.

\bibitem{boba91}
S.~P. Boyd and C.~H. Barratt.
\newblock {\em Linear controller design: limits of performance}.
\newblock 1991.

\bibitem{dahm04}
A.~De Brevern, S.~Hazout, and A.~Malpertuy.
\newblock Influence of microarrays experiments missing values on the stability
  of gene groups by hierarchical clustering.
\newblock {\em BMC bioinformatics}, 5(1):114, 2004.

\bibitem{buns86}
Angelika Bunse-Gerstner and Volker Mehrmann.
\newblock A symplectic qr like algorithm for the solution of the real algebraic
  riccati equation.
\newblock {\em Automatic Control, IEEE Transactions on}, 31(12):1104--1113,
  1986.

\bibitem{butl98}
R.~W. Butler.
\newblock Generalized inverse {G}aussian distributions and their {W}ishart
  connections.
\newblock {\em Scandinavian journal of statistics}, 25(1):69--75, 1998.

\bibitem{buwo03}
R.~W. Butler and A.~Wood.
\newblock Laplace approximation for bessel functions of matrix argument.
\newblock {\em Journal of Computational and Applied Mathematics},
  155(2):359--382, 2003.

\bibitem{byer87}
R.~Byers.
\newblock Solving the algebraic {R}iccati equation with the matrix sign
  function.
\newblock {\em Linear Algebra and its Applications}, 85:267--279, 1987.

\bibitem{eberlein95}
E.~Eberlein and U.~Keller.
\newblock Hyperbolic distributions in finance.
\newblock {\em Bernoulli}, pages 281--299, 1995.

\bibitem{herz55}
Carl~S Herz.
\newblock Bessel functions of matrix argument.
\newblock {\em Annals of Mathematics}, pages 474--523, 1955.

\bibitem{jorgensen82}
B.~J{\o}rgensen.
\newblock {\em Statistical properties of the generalized inverse {G}aussian
  distribution}.
\newblock Springer, 1982.

\bibitem{kong94}
A.~Kong, J~Liu, and W.H. Wong.
\newblock Sequential imputations and bayesian missing data problems.
\newblock {\em Journal of the American statistical association},
  89(425):278--288, 1994.

\bibitem{laub79}
Alan~J Laub.
\newblock A schur method for solving algebraic riccati equations.
\newblock {\em Automatic Control, IEEE Transactions on}, 24(6):913--921, 1979.

\bibitem{lawr05}
N.~Lawrence.
\newblock Probabilistic non-linear principal component analysis with {G}aussian
  process latent variable models.
\newblock {\em JMLR}, 6:1783--1816, 2005.

\bibitem{laur09}
N.~Lawrence and R.~Urtasun.
\newblock {Non-linear Matrix Factorization with {G}aussian Processes}.
\newblock In {\em ICML}, 2009.

\bibitem{lclw13}
T.~Li, E.~Chu, W.~Lin, and P.~Weng.
\newblock Solving large-scale continuous-time algebraic riccati equations by
  doubling.
\newblock {\em Journal of Computational and Applied Mathematics},
  237(1):373--383, 2013.

\bibitem{li13}
Y.~Li, M.~Yang, Z.~Qi, and Z.~Zhang.
\newblock Bayesian multi-task relationship learning with link structure.
\newblock In {\em ICDM}, pages 1115--1120. IEEE, 2013.

\bibitem{mackay03}
D.J.C. MacKay.
\newblock {\em Information theory, inference, and learning algorithms},
  volume~7.
\newblock Citeseer, 2003.

\bibitem{mink00}
T.~P. Minka.
\newblock Automatic choice of dimensionality for pca.
\newblock In {\em NIPS}, volume~13, pages 598--604, 2000.

\bibitem{owen13}
A.~Owen.
\newblock {\em Monte Carlo theory, methods and examples}.
\newblock 2013.

\bibitem{samn07}
R.~Salakhutdinov and A.~Mnih.
\newblock {Probabilistic Matrix Factorization}.
\newblock In {\em NIPS}, 2007.

\bibitem{samn08b}
R.~Salakhutdinov and A.~Mnih.
\newblock {Bayesian Probabilistic Matrix Factorization using Markov Chain Monte
  Carlo}.
\newblock In {\em ICML}, 2008.

\bibitem{seshadri03}
V~Seshadri.
\newblock Some properties of the matrix generalized inverse {G}aussian
  distribution.
\newblock {\em Statistical methods and practice. Recent advances. Narosa
  Publishing House, New Delhi}, pages 47--56, 2003.

\bibitem{seshadri08}
V~Seshadri and J~Weso{\l}owski.
\newblock More on connections between {W}ishart and matrix {GIG} distributions.
\newblock {\em Metrika}, 68(2):219--232, 2008.

\bibitem{smho72}
W.B. Smith and R.R. Hocking.
\newblock Algorithm as 53: {W}ishart variate generator.
\newblock {\em Applied Statistics}, 1972.

\bibitem{tibi99}
M.~E. Tipping and C.~M. Bishop.
\newblock Probabilistic principal component analysis.
\newblock {\em Journal of the Royal Statistical Society: Series B},
  61(3):611--622, 1999.

\bibitem{wish28}
John Wishart.
\newblock The generalised product moment distribution in samples from a normal
  multivariate population.
\newblock {\em Biometrika}, pages 32--52, 1928.

\bibitem{yalz13}
M.~Yang, Y.~Li, and Z.~Zhang.
\newblock Multi-task learning with {G}aussian matrix generalized inverse
  {G}aussian model.
\newblock In {\em ICML}, 2013.

\bibitem{yoshii13}
K.~Yoshii and R.~Tomioka.
\newblock Infinite positive semidefinite tensor factorization for source
  separation of mixture signals.
\newblock In {\em ICML}, 2013.

\end{thebibliography}

\end{document}